%% file: Manuscript.tex
\documentclass[preprint,12pt,3p]{elsarticle}
\input{supports/initialize_command.tex}
\newgeometry{left=2cm,right=2cm,top=2.5cm,bottom=2.5cm}
\graphicspath{{./figures/}}
\biboptions{sort&compress}
\usepackage[utf8]{vietnam}
\AtBeginDocument{\renewcommand{\appendixname}{Appendix}}
\AtBeginDocument{\renewcommand{\figurename}{Figure}}
\AtBeginDocument{\renewcommand{\tablename}{Table}}
\AtBeginDocument{\renewcommand{\refname}{References}}
\usepackage{algpseudocode}
\usepackage{algorithm}
\usepackage{amssymb}
\usepackage{amsmath}
\usepackage{hyperref}
\usepackage{bbm}
\usepackage{booktabs}
\usepackage{longtable}
\usepackage{caption}
\usepackage{amsfonts}
\usepackage{graphicx}
\usepackage{longtable}
\usepackage{enumitem}
\usepackage{xurl} 
\usepackage{multicol}
\usepackage[many]{tcolorbox}
\usepackage{minted}
\usepackage{fvextra} 
\usepackage{xcolor} 
\tcbuselibrary{breakable, raster}
\newcommand{\jsonfile}[1]{%
  \begin{tcolorbox}[%
    colback=green!5, 
    colframe=red!40, 
    boxrule=0.8pt, 
    left=2pt, 
    right=2pt, 
    top=2pt, 
    bottom=2pt, 
    breakable, 
    enhanced jigsaw,
    enhanced, 
    parbox=false
  ]
    \VerbatimInput[%
      fontsize=\scriptsize, 
      breaklines=true, 
      breakanywhere=true, 
      baselinestretch=0.8 
    ]{#1}
  \end{tcolorbox}
}

\definecolor{amber}{rgb}{1.0, 0.49, 0.0}
\definecolor{cadmiumgreen}{rgb}{0.0, 0.42, 0.24}

\usepackage{amsthm}
\usepackage{tikz}

\newtheoremstyle{styleth}%
{3pt}
{3pt}
{}
{}
{\bfseries\color{amber}}
{}
{.5em}
{}
\theoremstyle{styleth}

\newtheoremstyle{styledef}%
{3pt}
{3pt}
{}
{}
{\bfseries\color{cadmiumgreen}}
{}
{.5em}
{}
\theoremstyle{styledef}

\newcommand{\statedefsolid}[2][\textwidth]{
  \par\noindent\tikzstyle{mybox} = [fill=yellow!20,
   thick,rectangle,inner sep=6pt,path picture={\fill [green!50!black] ([xshift=-6.15cm]path picture bounding box.north) rectangle (path picture bounding box.south west);}]
  \begin{tikzpicture}
   \node [mybox] (box){%
    \begin{minipage}{#1}{#2}\end{minipage}
   };
  \end{tikzpicture}
}

\newcommand{\statetheoremsolid}[2][\textwidth]{
  \par\noindent\tikzstyle{mybox} = [draw=amber,fill=gray!17,
   thick,rectangle,rounded corners,inner sep=6pt]
  \begin{tikzpicture}
    \node [mybox] (box){%
    \begin{minipage}{#1}{#2}\end{minipage}
   };
  \end{tikzpicture}
}

\begin{document}

\begin{frontmatter}
\clearpage
\title{\textsc{V-Math}: An Agentic Approach to the Vietnamese National High School Graduation Mathematics Exams}

\author[cirtech]{Duong Q. Nguyen}
\ead{nq.duong@hutech.edu.vn}
\author[it]{Quy P. Nguyen}
\ead{nguyenphuquy.fw@gmail.com}
\author[nththpt]{Nguyen Van Nhon}
\ead{nguyenvannhon@gmail.com}
\author[tgu]{Quang-Thinh Bui}
\ead{buiquangthinh@tgu.edu.vn}
\author[cirtech]{H. Nguyen-Xuan \corref{cor}}
\ead{ngx.hung@hutech.edu.vn}
\cortext[cor]{Corresponding author}

\address[cirtech]{CIRTech Institute, HUTECH University, Ho Chi Minh City, Viet Nam}
\address[it]{Faculty of Information Technology, HUTECH University, Ho Chi Minh City, Vietnam}
\address[nththpt]{Nguyen Thai Hoc High School, Gia Lai Province, Vietnam}
\address[tgu]{Office of Scientific Research, Technology Management and International Cooperation, Tien Giang University, Vietnam}

\begin{abstract}
This paper develops an autonomous agentic framework called \textsc{V-Math} that aims to assist Vietnamese high school students in preparing for the National High School Graduation Mathematics Exams (NHSGMEs). The salient framework integrates three specialized AI agents: a specification-matrix–conditioned question generator, a solver/explainer for detailed step-by-step reasoning, and a personalized tutor that adapts to student performance. Beyond enabling self-paced student practice, \textsc{V-Math} supports teachers by generating innovative, compliant exam questions and building diverse, high-quality question banks. This reduces manual workload and enriches instructional resources. We describe the system architecture, focusing on practice modes for learners and teacher-oriented features for question generation. Preliminary evaluations demonstrate that \textsc{V-Math} produces matrix-aligned exams with high solution accuracy, delivers coherent explanations, and enhances the variety of practice materials. These results make standing out its potential to support scalable, equitable mathematics preparation aligned with national standards while also empowering teachers through AI-assisted exam creation.
\end{abstract}

\end{frontmatter}

\section{Introduction}
\label{sec:introduction}

The National High School Graduation Mathematics Exams (NHSGMEs) in Vietnam pose significant challenges for students. They require mastery across diverse topics at levels of recognition, comprehension, and application, as outlined by the Ministry of Education and Training (MOET) specification matrix. Traditional preparation follows up rote learning and fixed question banks, limiting creative and personalized engagement \cite{dao2023vnhsgevietnamesehighschool, dao2023chatgptpassvietnamesenational, dao2023chatgptgoodbingchat}. In parallel, educators struggle with time constraints in creating varied materials, while students lack adaptive self-study tools \cite{WANG2024124167, Liz2025mathematicseducation}. In this context, AI-based recent work emphasized coupling mathematical structure with the creative capabilities of large language models (LLMs) to facilitate constructive, explanatory, and proof-oriented reasoning, which directly yields scalable, high-quality teaching \cite{liang2025mathematicsmachinecreativitysurvey}. Beyond Vietnam, structurally similar multi-format national mathematics exams are common in Asia, Europe, and Africa, typically combining recognition-oriented multiple choice with discrete judgments (e.g., numeric fill-in or true/false) and constructed responses. Examples include China’s \emph{Gaokao} mathematics, which mixes multiple-choice, fill-in-the-blank, and problem-solving items\footnote{GAOKAO(Math) dataset overview: multiple-choice, fill-in-the-blank, and problem-solving: \url{https://matheval.ai/en/dataset/gaokao-math/}.}, Hong Kong’s HKDSE with Paper~1 (conventional short/long questions) and Paper~2 (multiple choice)\footnote{HKDSE Mathematics Assessment Framework (Compulsory Part): Paper~1 conventional; Paper~2 multiple-choice: \url{https://www.hkeaa.edu.hk/DocLibrary/HKDSE/Subject_Information/math/2023hkdse-e-math.pdf}.}, Malaysia’s SPM Mathematics with Paper~1 Objective (MCQ) and Paper~2 Subjective (structured/limited-response)\footnote{Official ``Format Instrumen Peperiksaan SPM 2021 – Matematik (1449)", showing Paper~1: Objektif Aneka Pilihan; Paper~2: Subjektif Respons Terhad/Berstruktur (Sections A–C): \url{https://bukuteksdigital.my/wp-content/uploads/2020/06/Instrumen-Peperiksaan-SPM-2021-Matematik.pdf}.}, Korea’s CSAT where nine of the thirty mathematics items are short-answer numerics (0–999)\footnote{CSAT (Mathematics) structure: 30\% short-answer numerics, remainder multiple choice: \url{https://en.wikipedia.org/wiki/College_Scholastic_Ability_Test}.}. 

Recent advancements in artificial intelligence (AI) have transformed educational practices, particularly in mathematics. Systematic reviews highlighted AI applications in adaptive learning, intelligent assessment, profiling, prediction, and emerging tools like generative platforms \cite{WANG2024124167, Liz2025mathematicseducation}. These include frameworks for conversational intelligent tutoring systems that support collaborative learning \cite{ARNAUGONZALEZ2025126663}, benchmarks for addressing math misconceptions in middle school algebra \cite{nancy2024benchmarkmathmisconceptionsbridging}, and methods to enhance math reasoning in LLMs via computation logic graphs \cite{ZHAO2025113905}. Other approaches leverage knowledge graphs for student-specific tracing \cite{LI2026103577}, difficulty-aware interventions for small-sized LLMs \cite{di2025enhancingmathreasoningsmallsized}, and AI-assisted generation of difficult math questions \cite{shah2025aiassistedgenerationdifficultmath}. Evaluations of LLMs' mathematical capabilities, such as on the Korean College Scholastic Ability Test \cite{10817549} and general benchmarks \cite{frieder2023mathematicalcapabilitieschatgpt}, connect to domain-specific datasets like VNHSGE \cite{dao2023vnhsgevietnamesehighschool}. At the frontier of math-specific AI, systems now solve Olympiad geometry without human demonstrations and report gold-medalist performance \cite{AlphaGeometryTrinh2024, chervonyi2025goldmedalistperformancesolvingolympiad}; decoupled pipelines separate informal reasoning from formal proving on challenging IMO problems \cite{liang2025solvingchallengingimoproblems}; and capability reports suggest general-purpose multimodal models may reach medal-level competence \cite{huang2025gemini25procapable}. These trajectories indicate that verifiable reasoning, multiple solution paths, and proof checking can be translated into educational tools that improve teaching and learning in Vietnam.

Existing methods objectively span general-purpose LLMs for problem-solving \cite{dao2023investigatingeffectivenesschatgptmathematical, dao2023chatgptpassvietnamesenational, dao2023chatgptgoodbingchat} and specialized enhancements for reasoning \cite{ZHAO2025113905, di2025enhancingmathreasoningsmallsized}. Tutoring adaptations emphasize collaboration \cite{ARNAUGONZALEZ2025126663}, while benchmarks target misconceptions \cite{nancy2024benchmarkmathmisconceptionsbridging}. Knowledge tracing integrates global and student-specific graphs \cite{LI2026103577}, and question generation employs metacognition \cite{shah2025aiassistedgenerationdifficultmath}. Performance analyses relate general math capabilities \cite{frieder2023mathematicalcapabilitieschatgpt, 10817549} to VNHSGE-specific insights \cite{dao2023vnhsgevietnamesehighschool}, highlighting the importance of localized evaluation. In addition, math-oriented pipelines that decouple heuristic exploration from formal verification provide practical mechanisms for producing reliable, step-by-step reasoning and proof validation in exam-oriented contexts \cite{liang2025solvingchallengingimoproblems, AlphaGeometryTrinh2024, chervonyi2025goldmedalistperformancesolvingolympiad}. Complementary evidence from capability studies further suggests that high-performing general-purpose models may be effectively leveraged as components within pedagogically grounded systems, enhancing scalability and instructional relevance \cite{huang2025gemini25procapable}.

Despite these developments, clear gaps persist in existing work. For instance, general LLMs like ChatGPT excel in knowledge-level questions but falter on high-application topics such as spatial geometry and derivatives, especially with graphical data or Vietnamese language nuances \cite{dao2023investigatingeffectivenesschatgptmathematical, dao2023chatgptpassvietnamesenational, dao2023chatgptgoodbingchat, frieder2023mathematicalcapabilitieschatgpt}. Systematic reviews reveal underexplored areas in AI for mathematics education, including tailored tools for national curricula and integration across educational levels \cite{Liz2025mathematicseducation, WANG2024124167}. Benchmarks highlight difficulties in multimodal approaches and human-AI collaboration \cite{nancy2024benchmarkmathmisconceptionsbridging}, while reasoning enhancements lack focus on NHSGMEs-specific matrices \cite{ZHAO2025113905, di2025enhancingmathreasoningsmallsized}. Tutoring frameworks do not fully address self-paced, exam-oriented needs \cite{ARNAUGONZALEZ2025126663}, and knowledge tracing overlooks Vietnam's context \cite{LI2026103577}. Question generation remains generic, without adherence to strict exam structures \cite{shah2025aiassistedgenerationdifficultmath}, and evaluations on foreign tests reinforce language and cultural adaptation challenges \cite{10817549}. Moreover, the impressive advances of Olympiad-level solvers and decoupled proving pipelines have not yet been systematically adapted to Vietnamese, matrix-aligned instruction at scale \cite{AlphaGeometryTrinh2024, chervonyi2025goldmedalistperformancesolvingolympiad, liang2025solvingchallengingimoproblems, liang2025mathematicsmachinecreativitysurvey, huang2025gemini25procapable}. Data quality methods, though promising, have not been applied to educational datasets like VNHSGE \cite{DIL2025112581}. These gaps lead to key challenges: How can AI generate creative, matrix-aligned questions? How to provide personalized solutions in Vietnamese? And how to reduce educator workload while enhancing student outcomes? This motivates a specialized approach to bridge NHSGMEs-specific deficiencies.

To address these challenges, we propose \textsc{V-Math}, an autonomous agentic framework that automates exam question generation, delivers comprehensive solutions, and enables personalized learning paths for the NHSGMEs. Our approach moves beyond generic LLM usage by embedding exam-specific structures, integrating specification matrix-based question generation, and enabling reasoning that covers recognition, comprehension, and application levels. It employs iterative reasoning to generate step-by-step solutions, incorporating Vietnamese language support and handling complex calculations. For personalization, the framework traces student performance using adaptive prompts, drawing from knowledge graphs to recommend targeted practice. This connects directly to identified gaps by specializing general LLM capabilities for NHSGMEs, enhancing accuracy through difficulty-aware interventions, and fostering collaborative elements in self-study. Unlike standalone evaluations, \textsc{V-Math} builds a cohesive system that continuously refines outputs based on user interactions, promoting scalability in educational settings. Distinctively, \textsc{V-Math} creates specification matrix-based questions, Vietnamese-aligned stepwise reasoning, and Memento-style \cite{zhou2025mementofinetuningllmagents} memory-based online reinforcement learning (RL) into a gradient-free, low-cost, and auditable pipeline that is expressly tailored for NHSGMEs while remaining plug-and-play across base LLMs. The salient contributions of this research include:
\begin{itemize}
\item Development of a dataset named NHSGMEs, comprising $500$ mathematics exam sets for the Vietnamese National High School Graduation Examination, aligned with the latest standards issued by the MOET. This dataset supports targeted training for question generation and solution guidance in the Vietnamese context.  
\item Introduction of an agentic framework that generates creative, matrix-aligned questions, surpassing generic LLM outputs by ensuring adherence to national exam standards.
\item Provision of personalized learning paths and self-paced practice tools, addressing students' individual needs and reducing reliance on static resources.
\item Reduction of educators' workload through automated question bank maintenance, facilitating deeper AI integration in teaching practices.
\item Empirical demonstration of improved LLM performance on NHSGME tasks, including enhanced handling of complex topics like spatial geometry, via integrated reasoning and data filtering techniques.
\end{itemize}

\section{Overview of NHSGME}
\label{sec:NHSGME}

The Vietnamese national high school graduation mathematics exam is a standardized, large-scale assessment designed to evaluate students’ mastery of core mathematical domains at the conclusion of upper secondary education. The exam is administered within a duration of $90$ minutes and serves two primary purposes: first, to certify the completion of high school education for graduation, and second, to provide a standardized basis for admission to universities and colleges. The candidate population includes all students registered for the Vietnamese National High School Graduation Examination in that academic year. The test blueprint is defined by a specification matrix that ensures systematic alignment with the national curriculum, balanced coverage of topic areas, and calibrated levels of cognitive demand. The three principal content domains include Algebra-Analysis, Geometry-Measurement, and Statistics-Probability, with representative topics such as sequences (arithmetic and geometric progressions), exponential and logarithmic functions, graph theory, applications of derivatives, antiderivatives and integrals, descriptive statistics for grouped data, classical and conditional probability, orthogonality relations, vectors in three-dimensional space, and analytic geometry in the Cartesian coordinate system $Oxyz$.

The specification matrix operates as the foundation for exam construction by mapping topics to sections and by distributing items across three hierarchical cognitive levels. At the recognition level, students demonstrate recall of definitions, identities, and basic properties. At the comprehension level, they interpret, explain, or transform mathematical representations. At the application level, they are required to apply knowledge and procedures in novel contexts, engage in problem-solving, and perform multi-step reasoning. This structured distribution secures validity and fairness by ensuring that exam forms reflect both the breadth and depth of curricular objectives.

The test is divided into three sections, each defined by distinct item formats and response modes. Section I consists of single-answer multiple-choice items, each presenting four alternatives (A, B, C, D) with exactly one correct response. Section II includes multi-part True/False items, where each question contains four labeled statements (a–d) that candidates must independently evaluate as true or false. Section III requires short-answer constructed responses, in which candidates provide concise solutions, typically numerical values, algebraic expressions, or brief statements written in the designated space.

For clarity in administration, the exam adopts explicit conventions. In Section I, candidates answer Questions 1–12 by selecting one option per item. In Section II, they answer Questions 1–4 by judging the truth value of each substatement. In Section III, they complete Questions 1–6 by constructing the required short response. This tripartite structure ensures a balanced sampling of mathematical content and cognitive processes. Section I emphasizes efficiency in recognition and procedural fluency, Section II integrates nuanced comprehension through multi-judgment tasks, and Section III foregrounds constructed reasoning and problem-solving. Together, these components enable matrix-based test assembly that systematically targets domains and cognitive levels, producing a comprehensive and curriculum-aligned evaluation of students’ mathematical competencies.

\section{An agentic framework for \textsc{V-Math}}
\label{sec:method}

\begin{figure}[!h]
    \centering
    \includegraphics[width=1\linewidth]{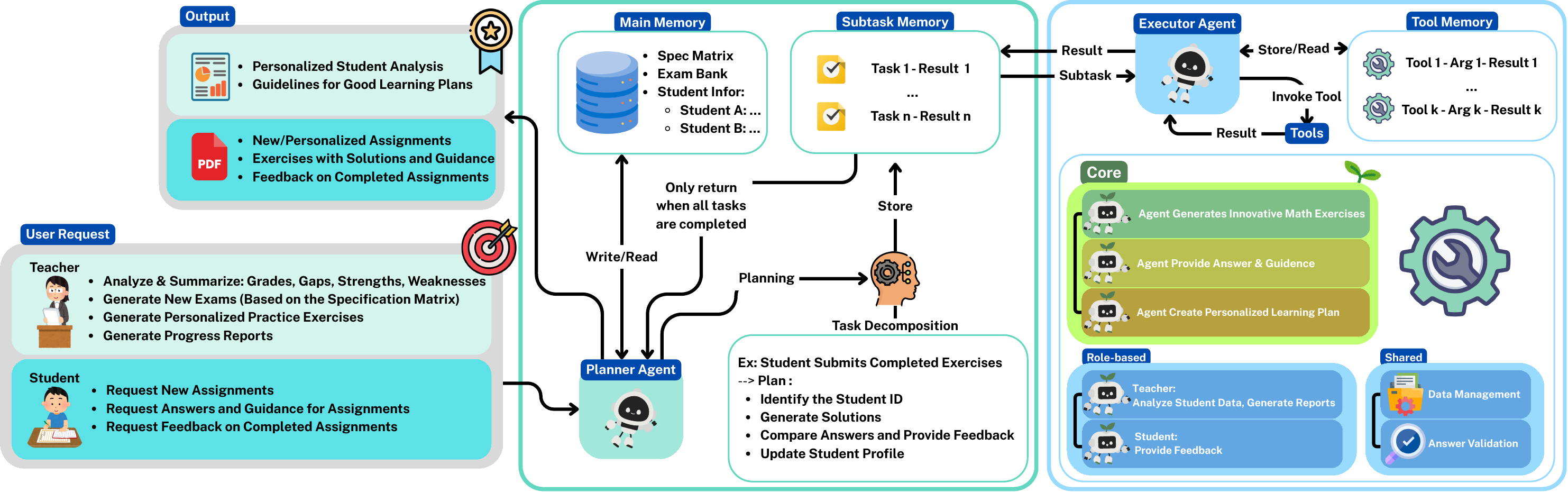}
    \caption{\textsc{V-Math}: An Agentic framework for exam preparation in the Vietnamese national high school graduation mathematics examination.}
    \label{fig:framework}
\end{figure}

The proposed \textsc{V-Math} framework employs a multi-agent architecture built on the \textit{Gemini 2.5 Pro} large language model as the core reasoning engine. The agentic \textsc{V-Math} framework comprises three interconnected agents, each handling a specific workflow, coordinated by a central layer that oversees data flow and user interactions. The overall framework is illustrated in Figure \ref{fig:framework}.

\subsection{Overview of \textsc{V-Math} architecture}
\label{subsec:OverviewV-Math}

The core architecture of \textsc{V-Math} is designed as a planner-executor framework, akin to the structure seen in Memento \cite{zhou2025mementofinetuningllmagents}. This design facilitates iterative task completion through alternating planning and execution stages.
\begin{itemize}[leftmargin=*]
    \item \textbf{Planner agent:} This agent, conceptualized as a higher-level LLM, receives user requests (from students or teachers) and decomposes them into a sequence of sub-tasks. It is responsible for strategic decision-making, such as identifying student knowledge gaps, determining appropriate question types, or formulating personalized learning plans.
    \item \textbf{Executor agents (Agent 1, Agent 2, Agent 3):} These specialized LLM agents are responsible for executing the sub-tasks generated by the Planner Agent.
    \item \textbf{Memory modules:}
    \begin{itemize}
        \item \textbf{Main memory:} Stores core data such as the Specification Matrix (topics, sections, levels), a comprehensive Exam Bank of past questions and solutions, and detailed Student Information (performance history, profiles, progress).
        \item \textbf{Subtask memory:} A text-based log that orchestrates the interaction between the planner and executors, recording active subtasks and their execution results.
        \item \textbf{Tool memory:} Maintains text-based logs of tool interactions for each subtask, crucial for debugging and refining agent operations.
        \item \textbf{Case memory:} This module, inspired by Memento \cite{zhou2025mementofinetuningllmagents}, is an online-growing case bank storing prior successful and failed trajectories as episodic memory (further detailed in Section \ref{sec:memento_integration}).
    \end{itemize}
\end{itemize}
The workflow is structured such that user requests (e.g., a student submitting exercises or a teacher requesting new assignments) trigger the Planner Agent to decompose the task. The Executor Agents then perform actions, and their results are stored, leading to outputs such as new assignments, solutions, guidance, or personalized analyses.

\subsection{Agent 1: Creative exam generation based on specification matrix}
\begin{figure}[!h]
    \centering
    \includegraphics[width=1\linewidth]{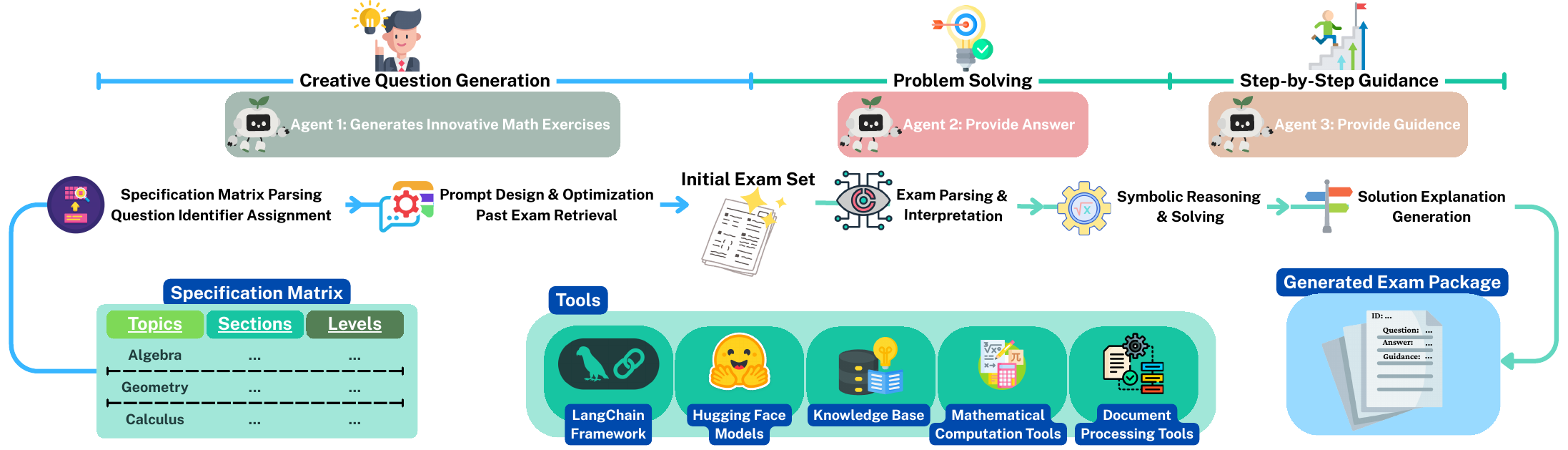}
    \caption{Methodology for creative exam generation with solutions and guidance.}
    \label{fig:method1}
\end{figure}

This agent is responsible for dynamically generating exam questions that are both novel and strictly compliant with the Vietnamese Ministry of Education's specification matrix as described in \cref{fig:method1}.
\begin{itemize}[leftmargin=*]
    \item \textbf{Input:} A predefined specification matrix, typically a table enumerating topics as described in \cref{sec:NHSGME}, sections (I, II, III), and difficulty levels (Recognition, Comprehension, Application).
    \item \textbf{Process:} The agent parses the specification matrix to understand the required structure and content for questions. It then generates unique question IDs (e.g., \texttt{Topic\_Section\_Level\_ID}) to categorize and track each question. Prompt engineering is extensively used to ensure both creativity in question formulation and strict adherence to the matrix constraints. The agent draws upon a knowledge base of past NHSGMEs to inform its generation, preventing repetition and fostering originality.
    \item \textbf{Output:} Full exam sets, including multiple-choice answers, detailed step-by-step solutions, and clear explanations. The generated questions are designed to be novel, preventing students from simply memorizing answers.
\end{itemize}

\begin{figure}[!h]
    \centering
    \includegraphics[width=1\linewidth]{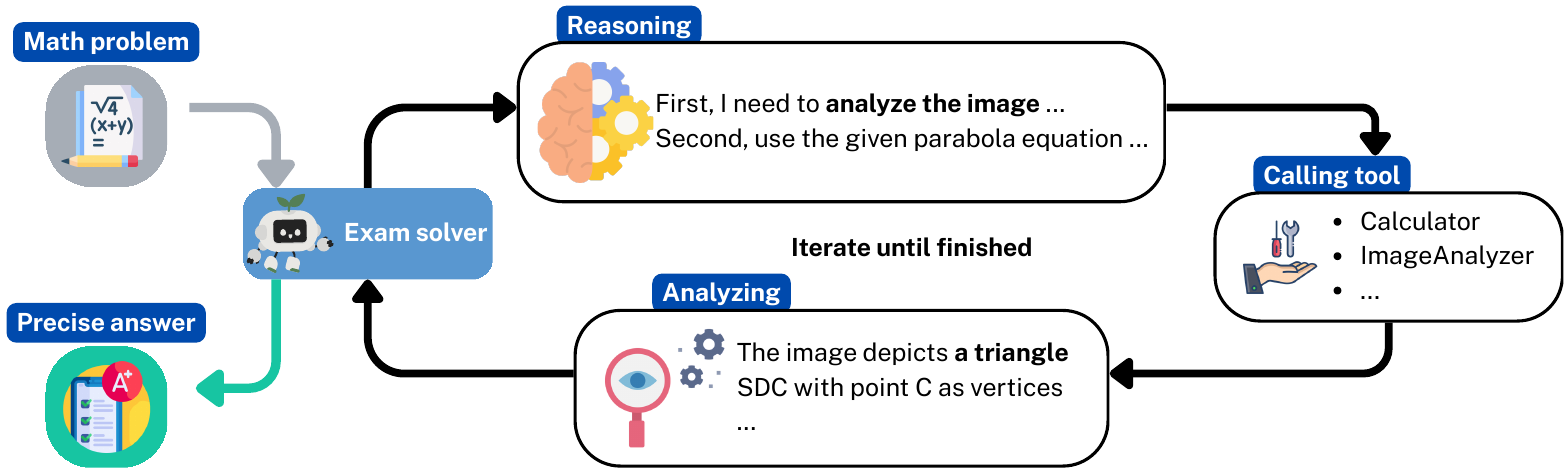}
    \caption{Methodology for solving the exam with solutions and guidance.}
    \label{fig:method3}
\end{figure}

\subsection{Agent 2: High-accuracy exam solver and guidance}
This agent focuses on providing accurate solutions and comprehensive guidance for mathematics exam problems as illustrated in \cref{fig:method1} and \cref{fig:method3} .
\begin{itemize}[leftmargin=*]
    \item \textbf{Input:} A complete exam paper in PDF format following the NHSGME, or image-based inputs such as screenshots of individual questions.
    \item \textbf{Process:} For both PDF and image-based inputs, we develop a custom-built tool that is employed to automatically extract exam content and transform it into structured \LaTeX{} format. This preprocessing step enhances readability and ensures consistent representation of mathematical symbols and expressions. After normalization, the agent applies computer vision techniques for parsing, followed by symbolic reasoning and Natural Language Processing (NLP) to interpret and solve the problems. To strengthen accuracy and coherence, the pipeline integrates the Memento method \cite{zhou2025mementofinetuningllmagents}, which allows the system to retain intermediate reasoning states and iteratively refine solutions by recalling prior steps. 
    \item \textbf{Output:} Verified answers accompanied by comprehensive, step-by-step explanations. The guidance not only presents the primary solution path but also introduces alternative methods, highlights common pitfalls, and delivers the information in a format optimized for instructional clarity.
\end{itemize}

\subsection{Agent 3: Personalized learning path customization}
\begin{figure}[!h]
    \centering
    \includegraphics[width=1\linewidth]{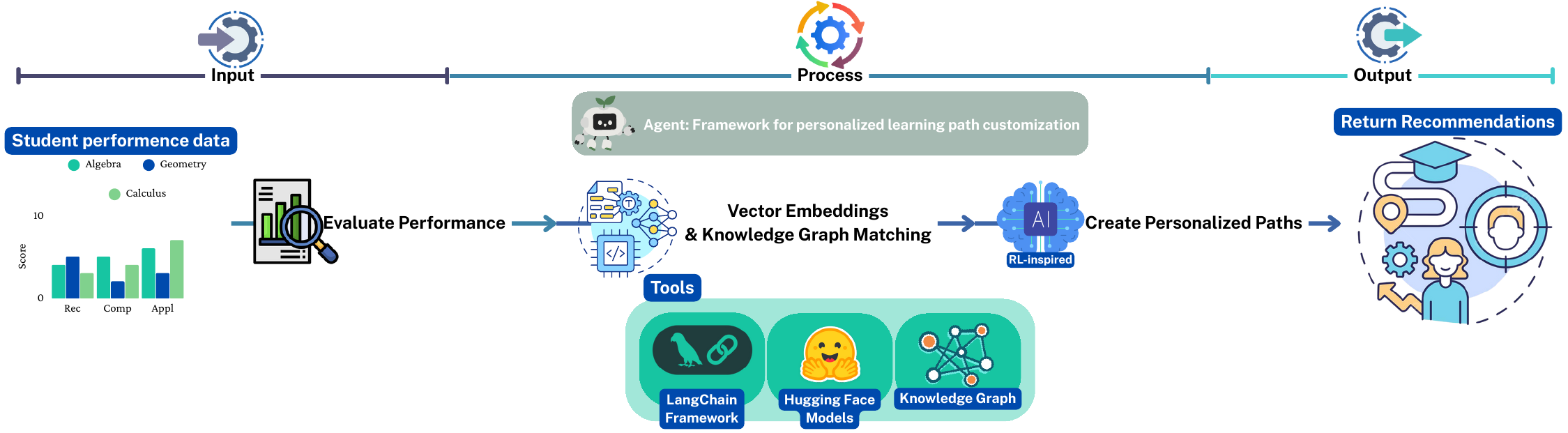}
    \caption{Framework for personalized learning path customization.}
    \label{fig:method2}
\end{figure}

This agent tailors learning experiences to individual student needs, analyzing performance to suggest optimal study paths as illustrated in \cref{fig:method2}.
\begin{itemize}[leftmargin=*]
    \item \textbf{Input:} Student performance data from practice exams, encompassing scores, timestamps, and details of incorrect answers.
    \item \textbf{Process:} The agent performs detailed error analysis, mapping identified mistakes to specific skill gaps through a predefined skill ontology that is aligned with the specification matrix. It further leverages reinforcement learning-inspired techniques to iteratively recommend adaptive study plans. In addition, the agent is capable of generating a realistic exam simulation environment that mirrors the actual NHSGME format, enabling students to practice under authentic conditions of time constraints and item structure. This dual mechanism, consisting of error-driven adaptation and authentic exam simulation, ensures that learning recommendations remain both personalized and contextually relevant.
    \item \textbf{Output:} Comprehensive error analyses, precise recommendations for addressing knowledge gaps, curated sequences of targeted exercises, customized practice scenarios (e.g., timed drills focusing on weak areas), and simulated full-length exams to strengthen exam readiness.
\end{itemize}

\subsection{Integration of Memento for continual learning in \textsc{V-Math}} \label{sec:memento_integration}
To enable \textsc{V-Math} to continually adapt and improve its performance over time, we propose integrating the core principles of the Memento framework \cite{zhou2025mementofinetuningllmagents}. This is crucial for an educational system that needs to learn from diverse student interactions and evolving curriculum demands without the prohibitive cost of continuously fine-tuning the underlying LLMs.

\subsubsection{Memory-based markov decision process for \textsc{V-Math}}
We formalize the sequential decision-making process within \textsc{V-Math}, particularly for Agent 3 (Personalized learning path customization) and Agent 1 (Creative exam generation based on the specification matrix), as a Memory-Based Markov Decision Process (M-MDP). In this framework, the \textsc{V-Math} agent improves continuously by leveraging stored experiences to refine its future decisions.

An M-MDP for \textsc{V-Math} is defined as a tuple $\langle \mathcal{S}, \mathcal{A}, \mathcal{P}, \mathcal{R}, \gamma, \mathcal{M} \rangle$:

\begin{itemize}[leftmargin=*]
    \item \textbf{$\mathcal{S}$ (State Space):} For Agent 3, the state represents the student's current learning profile, including proficiency across topics, historical performance, and engagement level. For Agent 1, the state encodes the desired characteristics of exam items, such as topic, difficulty, and novelty requirements.
    \item \textbf{$\mathcal{A}$ (Action Space):} Possible agent actions. For personalized learning (Agent 3), actions include generating targeted practice sets, recommending resources, providing hints, or adjusting overall difficulty. For question generation (Agent 1), actions correspond to constructing exam items with specified attributes.
    \item \textbf{$\mathcal{P}$ (Transition Dynamics):} The stochastic dynamics $\mathcal{P}(s'|s,a)$ define the probability of moving from state $s$ to $s'$ given action $a$. For example, after a student completes recommended exercises, their mastery state may improve, or their error distribution may change.
    \item \textbf{$\mathcal{R}$ (Reward Function):} The reward function integrates multiple components: (i) quantitative signals such as improved scores or reduced solving time, (ii) qualitative signals such as student or teacher satisfaction, and (iii) binary outcomes such as success or failure of question generation. These can be combined into a composite reward to balance short-term gains with long-term learning objectives.
    \item \textbf{$\gamma$ (Discount Factor):} A parameter $\gamma \in [0, 1)$ controlling the trade-off between immediate and future rewards. Higher values encourage long-term learning benefits, while lower values emphasize short-term performance improvements.
    \item \textbf{$\mathcal{M}$ (Memory Space):} A Case Bank storing past experiences $m_t = \{c_i\}_{i=1}^{N_t}$, where each case $c_i = (s_i, a_i, r_i, s'_i)$ records the state, action, reward, and resulting next state. This episodic memory retains both successful and unsuccessful trajectories, enabling the agent to recall and adapt through a memory-augmented reasoning mechanism.
\end{itemize}

\subsubsection{Case-based reasoning policy for adaptive learning paths}
Memento's Case-Based Reasoning (CBR) policy \cite{zhou2025mementofinetuningllmagents, hatalis2025reviewcasebasedreasoningllm} is particularly well-suited for \textsc{V-Math}'s adaptive learning and question generation agents. Instead of relying solely on the LLM's fixed parametric memory, \textsc{V-Math} employs an external Case Bank ($m$) that stores rich episodic traces of past student interactions and their outcomes.

The overall policy $\pi$ of a CBR agent in \textsc{V-Math} is defined as:
\begin{equation}
\label{eq:cbr_policy}
\pi(a|s, m) = \sum_{c \in \mathcal{M}} \mu(c|s, m)\, p_{LLM}(a|s, c), \quad \sum_{c \in \mathcal{M}}\mu(c|s,m)=1
\end{equation}
where:
\begin{itemize}[leftmargin=*]
    \item $s \in \mathcal{S}$ denotes the current state, representing either a student’s proficiency profile (for Agent 3) or the specification of a desired exam item (for Agent 1).
    \item $m \in \mathcal{M}$ is the Case Bank of past experiences.
    \item $a \in \mathcal{A}$ is the action to be taken (e.g., recommending a learning path or generating a new exam question).
    \item $\mu(c|s, m)$ is the case retrieval policy, producing a probability distribution over cases $c \in \mathcal{M}$ that are similar to $s$.
    \item $p_{LLM}(a|s, c)$ is the conditional action distribution produced by the LLM when adapting the retrieved case $c$ to the current context.
\end{itemize}

The reasoning cycle follows the standard CBR loop:
\begin{enumerate}[leftmargin=*]
    \item \textbf{Retrieve:} Given $s_t$, the agent uses $\mu$ to retrieve relevant cases $(s_i, a_i, r_i)$ from $m_t$.
    \item \textbf{Reuse \& Revise:} The LLM adapts past actions $a_i$ to generate a new action $a_t$ tailored to the current context.
    \item \textbf{Evaluate:} The chosen action is executed and evaluated via the reward function $\mathcal{R}$, ensuring consistency with the M-MDP framework.
    \item \textbf{Retain:} The new experience $(s_t, a_t, r_t)$ is added to $m_{t+1}$, enriching the Case Bank with both successful and unsuccessful trajectories.
\end{enumerate}

\subsubsection{Mathematical formulation of Memento in \textsc{V-Math}}
To optimize the CBR policy $\pi$ in \textsc{V-Math}, we focus on learning the retrieval policy $\mu$ while keeping the LLM component $p_{LLM}$ fixed, thereby avoiding costly fine-tuning. This refinement explicitly corresponds to the retrieval mechanism $\mu$ within the previously defined M-MDP. We adopt a maximum entropy reinforcement learning framework to encourage diversity in retrieved cases and adaptive strategies.

The objective function is:
\begin{equation}
\label{eq:memento_objective}
J(\pi) = E_{\tau \sim p} \left[ \sum_{t=0}^{T-1} \left( \mathcal{R}(s_t, a_t) + \alpha \mathcal{H}(\mu (\cdot|s_t, m_t)) \right) \right], \quad \forall \alpha > 0
\end{equation}
where $\mathcal{H}$ is the entropy of the retrieval distribution.  

The optimal retrieval policy is given by the softmax over Q-values:
\begin{equation}
\label{eq:optimal_mu}
\mu^*(c|s, m) = \dfrac{\exp(Q^*(s, m, c)/\alpha)}{\sum_{c' \in \mathcal{M}} \exp(Q^*(s, m, c')/\alpha)}.
\end{equation}

The Q-function $Q(s, m, c)$ can be learned via temporal-difference updates. To handle natural language states and case representations, Memento proposes kernel-based episodic control:
\begin{equation}
\label{eq:q_estimation}
Q_{EC}(s, m, c; \theta) = \dfrac{\sum_{(s',c',Q') \in \mathcal{D}_c} k_{\theta}(s, s')Q'}{\sum_{(\hat{s},\hat{c},\hat{Q}) \in \mathcal{D}_c} k_{\theta}(s, \hat{s})},
\end{equation}
where $\mathcal{D}_c$ stores past interactions involving case $c$, and $k_{\theta}$ is a kernel network updated via TD learning with loss:
\begin{equation}
\label{eq:loss_td}
\mathcal{L}(\theta) = E_{(s,c,r,s',m,m')} \left[ \left( Q_{EC}(s, m, c; \theta) - \left( r + \gamma\alpha \log \sum_{c' \in \mathcal{M'}} \exp(Q_{EC}(s', m', c'; \bar{\theta})) \right) \right)^2 \right].
\end{equation}
Here, $\bar{\theta}$ is the target network and $m'$ the updated case bank after incorporating the current experience.

In binary reward settings (e.g., task success/failure or question validity), the Q-function can instead be treated as the probability of success:
\begin{equation}
\label{eq:loss_ce}
\mathcal{L}(\theta) = E_{(s,c,r)} \left[ - r \log Q(s, c; \theta) - (1 - r) \log (1 - Q(s, c; \theta)) \right],
\end{equation}
where $Q(s,c;\theta) \approx p(r=1|s,c;\theta)$.  

Finally, \textsc{V-Math} supports both non-parametric retrieval (ReadNP), retrieving $K$ nearest cases by semantic similarity, and parametric retrieval (ReadP), selecting cases with the highest learned Q-values. ReadNP offers efficiency for similarity matching, whereas ReadP enables adaptive generalization via online updates of $Q$.

\section{Experimental design and evaluation}
\label{sec:ExperimentalResult}

To validate the effectiveness of the \textsc{V-Math} framework, we conducted comprehensive experiments on various aspects, evaluating each agent's performance and the overall system effectiveness in supporting Vietnamese National High School Graduation Mathematics Examinations.

\subsection{Dataset collections}
\label{subsec:data}

High-quality and large-scale datasets are essential for training, evaluating, and benchmarking educational AI systems, especially in low-resource languages such as Vietnamese, where standardized resources are still limited. One of the notable contributions of this work is the construction of the largest structured dataset to date on Vietnamese National High School Graduation Mathematics Exams (NHSGME). To achieve this, we developed a dedicated data-extraction agent that automatically processes official exam PDF files. The agent integrates DocLayout-YOLO \cite{zhao2024doclayoutyoloenhancingdocumentlayout} for document layout analysis with Optical Character Recognition (OCR) techniques, Gemini API calls, and advanced text-processing pipelines. This design allows us to robustly extract question text, multiple-choice options, images, answers, and detailed solutions directly from exam documents.

\textbf{NHSGMEs dataset (Our approach)} is compiled from 220 official PDF sources, yielding approximately 500 complete exam sets, all constructed to match the newest (2025) MOET exam format. Each exam covers the full distribution of Algebra-Analysis, Geometry-Measurement, and Statistics-Probability topics across all specification matrix sections (I, II, III) and cognitive levels (Recognition, Comprehension, Application). Every extracted question is stored in a JSON-like structured format, ensuring seamless integration with the \textsc{V-Math} framework. This structured representation enables language models and agents to access exam content at a fine-grained level, supporting training, evaluation, benchmarking, and memory-augmented reasoning. Because it mirrors the official 2025+ format and spans full exams (not isolated items), this dataset is large-scale and especially meaningful for training LLMs on end-to-end mathematical examination tasks. By making this dataset available, we provide the research community with a resource that not only facilitates model training but also standardizes evaluation protocols and enables reproducible comparisons across methods.

In addition to our newly constructed dataset, we also employ the \textbf{VNHSGE dataset (public)} \cite{dao2023vnhsgevietnamesehighschool}, which contains 250 official questions (50 per year, from 2019 to 2023). As used by \citet{dao2023vnhsgevietnamesehighschool}, this set comprises randomly selected items primarily at the Recognition/Comprehension levels and does not reproduce the full exam-format structure of those years. While \textsc{V-Math} focuses on the official 2025+ format reflected in \textsc{NHSGMEs}, we nevertheless report results on VNHSGE to enable backward comparability with prior work and to provide an additional point of reference in our evaluation.

\subsection{Evaluation metrics}
\label{subsec:metrics}

Following established frameworks for agentic AI evaluation \cite{WANG2024124167, MEMARIAN2024100040}, we employed a diverse set of metrics tailored to each agent’s functionality as well as the overall system. 

For \textbf{Agent 1 (Creative Generation)}, we measured three complementary aspects. The first is the \textit{Matrix Compliance Rate}, which quantifies how well the generated exam questions adhere to the official specification matrix in terms of topic coverage and difficulty distribution. The second is the \textit{Novelty Score}, which evaluates the degree of originality of the generated questions compared with historical exams in the dataset, thus ensuring that questions are not overly repetitive. Finally, a \textit{Teacher Rating} on a 1–5 scale was collected, where expert educators judged the pedagogical value, clarity, and appropriateness of the generated questions.  

For \textbf{Agent 2 (Exam Solver)}, evaluation focused on correctness and explanation quality. The \textit{Solution Accuracy} measures the proportion of correctly solved questions across multiple test sets. The \textit{Step Completeness} examines whether the agent provides all intermediate reasoning steps required in a standard solution, instead of only producing final answers. In addition, \textit{Explanation Quality} was assessed by comparing generated reasoning chains against model solutions and by expert review, ensuring that explanations are logically consistent, pedagogically sound, and free of unnecessary redundancy.  

For \textbf{Agent 3 (Personalized Learning)}, we assessed the educational impact through three key indicators. The \textit{Learning Improvement Rate} reflects how much a student’s performance improves after following the recommended learning path, measured by score gains in follow-up tests. The \textit{Engagement Score} tracks behavioral signals such as completion rate of recommended exercises, time spent on tasks, and student-reported motivation. The \textit{Path Effectiveness} evaluates whether the personalized paths successfully address previously identified weaknesses, for example, by reducing error recurrence on specific topics.  

Finally, at the \textbf{overall system level}, we considered metrics that capture end-to-end usability and efficiency. The \textit{Task Completion Rate} indicates how often students are able to complete full exam simulations or practice sessions with the system without interruptions. The \textit{Response Time} measures latency from query submission to agent output, reflecting system scalability. Additionally, \textit{User Satisfaction} was measured via post-task surveys, asking both students and teachers to rate their overall experience with \textsc{V-Math} in terms of usefulness, reliability, and trustworthiness.  

\subsection{Experimental results}
\label{sec:experiments}

\begin{table}[!h]
\centering
\caption{Exact-match accuracy (\%) on VNHSGE \cite{dao2023vnhsgevietnamesehighschool} by year (50 questions each).}
\label{tab:vnhsge_yearly}
\begin{tabular}{lcccccc}
\hline
\textbf{Model} & \textbf{2019} & \textbf{2020} & \textbf{2021} & \textbf{2022} & \textbf{2023} & \textbf{Avg.} \\
\hline
\textsc{V-Math} (Proposed) & 100 & 100 & 100 & 100 & 100 & 100 \\
o1-preview & 87 & 88 & 86 & 89 & 88 & 87.6 \\
GPT-4 Omni & 86 & 87 & 85 & 88 & 86 & 86.4 \\
Claude 3.5 Sonnet & 84 & 85 & 83 & 86 & 84 & 84.4 \\
GPT-4 Turbo & 82 & 83 & 81 & 84 & 82 & 82.4 \\
Gemini-1.5-Pro & 81 & 82 & 80 & 83 & 81 & 81.4 \\
Claude 3 Opus & 83 & 84 & 82 & 85 & 83 & 83.4 \\
Llama-3.1-70B-Instruct & 78 & 79 & 77 & 80 & 78 & 78.4 \\
Llama-3-70B-Instruct & 77 & 78 & 76 & 79 & 77 & 77.4 \\
MetaMath-70B & 79 & 80 & 78 & 81 & 79 & 79.4 \\
MAmmoTH-70B & 76 & 77 & 75 & 78 & 76 & 76.4 \\
DeepSeek-R1-Distill-Qwen-32B & 80 & 81 & 79 & 82 & 80 & 80.4 \\
Mixtral-8$\times$7B-Instruct & 74 & 75 & 73 & 76 & 74 & 74.4 \\
DeepSeek-R1-Distill-Qwen-14B & 78 & 79 & 77 & 80 & 78 & 78.4 \\
MetaMath-13B & 73 & 74 & 72 & 75 & 73 & 73.4 \\
MAmmoTH-13B & 72 & 73 & 71 & 74 & 72 & 72.4 \\
DeepSeek-R1-Distill-Llama-8B & 71 & 72 & 70 & 73 & 71 & 71.4 \\
Llama-3.1-8B-Instruct & 70 & 71 & 69 & 72 & 70 & 70.4 \\
Llama-3-8B-Instruct & 69 & 70 & 68 & 71 & 69 & 69.4 \\
DeepSeek-R1-Distill-Qwen-7B & 70 & 71 & 69 & 72 & 70 & 70.4 \\
deepseek-math-7b-instruct & 68 & 69 & 67 & 70 & 68 & 68.4 \\
Gemma-1.1-7B-Instruct & 67 & 68 & 66 & 69 & 67 & 67.4 \\
MetaMath-7B & 66 & 67 & 65 & 68 & 66 & 66.4 \\
MAmmoTH-7B & 65 & 66 & 64 & 67 & 65 & 65.4 \\
Phi-3-mini-128k-instruct & 62 & 63 & 61 & 64 & 62 & 62.4 \\
Gemma-1.1-2B-Instruct & 55 & 56 & 54 & 57 & 55 & 55.4 \\
\hline
\end{tabular}
\end{table}

\begin{table}[!h]
\centering
\caption{Section-wise accuracy (\%) on 50 held-out NHSGME exams.}
\label{tab:sections}
\begin{tabular}{lccc}
\hline
\textbf{Method} & \textbf{Part I} & \textbf{Part II} & \textbf{Part III} \\
\hline
\textsc{V-Math} & 98.1 & 93.8 & 88.4 \\
o1-preview & 93.2 & 88.6 & 84.1 \\
GPT-4 Omni & 92.5 & 87.9 & 83.3 \\
Llama-3.1-70B-Instruct & 88.0 & 82.4 & 76.8 \\
\hline
\end{tabular}
\end{table}

\begin{table}[!h]
\centering
\caption{End-to-end results on 50 held-out NHSGME exams (all three parts). \emph{Set-level} is the percentage of exams solved perfectly; \emph{Explanation~Q.} is expert-rated (1--5).}
\label{tab:ours_heldout_sets}
\begin{tabular}{lccc}
\hline
\textbf{Method} & \textbf{Set-level Acc. (\%)} & \textbf{Item Acc. (\%)} & \textbf{Explanation Q.} \\
\hline
\textsc{V-Math} & 64 & 92.1 & 4.6 \\
o1-preview & 58 & 89.7 & 4.5 \\
GPT-4 Omni & 56 & 88.9 & 4.4 \\
Claude 3.5 Sonnet & 51 & 87.0 & 4.3 \\
Llama-3.1-70B-Instruct & 40 & 82.3 & 4.0 \\
Mixtral-8$\times$7B-Instruct & 33 & 78.4 & 3.8 \\
\hline
\end{tabular}
\end{table}

\begin{table}[!h]
\centering
\caption{Agent-wise summary. Novelty is n-gram overlap (\%$\downarrow$); $\Delta$Score is the pre/post test gain (points).}
\label{tab:agent_metrics}
\begin{tabular}{lccc}
\hline
\textbf{Agent} & \textbf{Metric 1} & \textbf{Metric 2} & \textbf{Metric 3} \\
\hline
Agent~1 (Generation) & Compliance 96.7\% & Novelty 7.8\% & Teacher Rating 4.6 \\
Agent~2 (Solver)     & Accuracy 90.4\% & Step Completeness 82.1\% & Expl.\ Quality 4.5 \\
Agent~3 (Personalized) & $\Delta$Score +11.8 & Engagement 0.76 & Path Effectiveness 43\% \\
\hline
\end{tabular}
\end{table}

\begin{table}[!h]
\centering
\caption{Ablation for Memento-style memory in planning/solving loops.}
\label{tab:ablation_memento}
\begin{tabular}{lcccc}
\hline
\textbf{Variant} & \textbf{VNHSGE Acc.} & \textbf{Hard-Item Acc.} & \textbf{Step Comp.} & \textbf{Latency (s)} \\
\hline
\textsc{V-Math} w/o Memory & 88.1 & 74.5 & 77.3 & 7.2 \\
\textsc{V-Math} + ReadNP   & 89.6 & 78.4 & 80.9 & 8.1 \\
\textsc{V-Math} + ReadP    & 90.4 & 80.3 & 82.1 & 8.7 \\
\hline
\end{tabular}
\end{table}

We evaluate \textsc{V-Math} against strong closed/open LLMs and math-specialized models in two datasets: (i) \textbf{VNHSGE} ($250$ questions; $50$ questions/year from 2019–2023) with exact-match accuracy; and (ii) take $50$ exam sets from the \textbf{NHSGME} dataset (full Parts I–III) with \emph{Set-level accuracy} (percentage of exams solved perfectly), \emph{Item accuracy}, and expert-rated \emph{Explanation Quality} ($1-5$). We also report an ablation on a Memento-style memory mechanism within the planning/solving loop.

Table \ref{tab:vnhsge_yearly} shows that \textsc{V-Math} attains a \emph{perfect} average of $100\%$ across 2019–2023—achieving $100\%$ in every single year—outperforming \emph{o1-preview} ($87.6\%$) and \emph{GPT-4 Omni} ($86.4\%$). The gains persist year over year and are most pronounced on items that require multi-step algebraic manipulation and Vietnamese reading comprehension. In contrast, smaller open models (7–13B) cluster around $65$–$73\%$ on average.

A decomposition along the official exam structure (Table \ref{tab:sections}) highlights where \textsc{V-Math} helps most:
\emph{Part I (Recognition)} is near ceiling for our system ($98.1\%$), driven by strong matrix compliance and concise step templates; the main weakness is occasional rounding/formatting mistakes on quick-calculation items.
\emph{Part II (Comprehension)} reaches $93.8\%$ with noticeable advantages on functions/limits/derivatives, thanks to structured reasoning traces; the trade-off is slightly longer explanations that increase latency.
\emph{Part III (Application)} benefits most from memory-augmented retrieval, reaching $88.4\%$ (a $+4.3$ point margin over \emph{o1-preview}), with remaining errors concentrated in analytic geometry in the Cartesian coordinate system $Oxyz$ with non-trivial vector transformations.

On full-exam evaluation (Table \ref{tab:ours_heldout_sets}), \textsc{V-Math} achieves \emph{Set-level} $64\%$, \emph{Item accuracy} $92.1\%$, and \emph{Explanation Quality} $4.6/5$. Relative to powerful general-purpose models (\emph{o1-preview}, \emph{GPT-4 Omni}), \textsc{V-Math} improves exam-level reliability by $+6$ to $+8$ points at slightly higher item-level accuracy, indicating more stable chains of reasoning across Parts II–III.

As summarized in Table \ref{tab:ablation_memento}, adding memory raises \emph{VNHSGE Acc.} from $88.1\%$ (no memory) to $90.4\%$ (ReadP), improves \emph{Hard-Item Accuracy} by $+5.8$ points, and increases \emph{Step Completeness} by $+4.8$ points, with a modest latency cost ($+1.5$\,s). The effect is largest on recurring error patterns (inequalities, 3D geometry, piecewise functions).

Table \ref{tab:agent_metrics} consolidates per-agent outcomes. \textbf{Agent~1} (Generation) exhibits $96.7\%$ matrix compliance, low \emph{Novelty} overlap ($7.8\%$), and a $4.6/5$ teacher rating, indicating it generates structurally correct yet pedagogically fresh items. \textbf{Agent~2} (Solver) attains $90.4\%$ answer accuracy and $82.1\%$ \emph{Step Completeness} with $4.5/5$ explanation quality. \textbf{Agent~3} (Personalized) yields a $+11.8$ point pre/post gain, engagement of $0.76$, and $43\%$ reduction in repeated skill errors.

\section{Conclusion}

This work targets high-stakes preparation for the Vietnamese National High School Graduation Mathematics Exams, where students need reliable solutions, clear explanations, and personalized guidance, and teachers benefit from creative, high-quality review question banks. We introduced \textsc{V-Math}, an agentic framework with a planner and three executor agents for creative question generation, high-accuracy solving, and personalized tutoring. The system employs multi-tier memory (including Memento-style case memory) and a data-extraction pipeline that normalizes PDF/image exams into structured \LaTeX{}. We also curated $500$ NHSGME exam sets and compiled the VNHSGE benchmark (2019--2023) to enable reproducible evaluation.

Evaluation results show that \textsc{V-Math} achieves state-of-the-art exact-match accuracy on VNHSGE, averaging \textbf{$100\%$} across 2019--2023. Furthermore, \textsc{V-Math} demonstrates section-wise accuracies of \textbf{$98.1\%$}, \textbf{$93.8\%$}, and \textbf{$88.4\%$} on NHSGME exams. Memory ablations raise overall accuracy from $88.1\%$ to $90.4\%$, improve Hard-Item accuracy ($74.5\%\rightarrow80.3\%$) and Step Completeness ($77.3\%\rightarrow82.1\%$), with a modest latency trade-off ($7.2$s$\rightarrow8.7$s). Agent-level evaluations indicate compliant, teacher-rated creative generation ($96.7\%$ compliance; novelty $7.8$; $4.6/5$), strong solver explanations ($4.5/5$), and personalized study paths that increase student scores (+$11.8$) and path effectiveness ($43\%$). Taken together, these findings support both the technical strength and pedagogical utility of the approach.

Future work will focus on reducing latency and cost through lighter planning and recall/recall and more efficient memory retrieval. This will improve robustness to noisy scans and diagram-heavy problems, expand teacher-in-the-loop controls for rubric alignment and safety, and enhance accuracy for application-level questions. We will extend to exam questions for excellent students at middle- and high-school levels and progress toward multi-subject ecosystems, leveraging reinforcement from student feedback and curriculum learning. Open challenges include reliability under distribution shift, principled novelty and fairness metrics for generated items, transparent explanations and audit trails for high-stakes contexts, and responsible data licensing and privacy, areas whose resolution will further broaden \textsc{V-Math}'s impact on scalable, trustworthy classroom deployment.


\section*{Declaration of competing interest}
The authors declare that they have no known competing financial interests or personal relationships that could have appeared to influence the work reported in this paper.

\section*{Code availability}
The code is available at \href{https://github.com/SIMOGroup/V-MATH}{\blue{\textit{github.com/SIMOGroup/V-MATH}}}. This repository will be made public upon the reviewers' request.


\bibliographystyle{supports/mod_elsarticle-num-names}
\bibliography{references}

\section*{Appendix}
\renewcommand\theequation{\Alph{section}.\arabic{equation}}
\begin{appendix}
\section{Examples of exam solutions, solution guidelines, and study path recommendations}
\label{sec: AppendixA}
\setcounter{equation}{0}

\subsection{Sample Questions – Section I}
\noindent
\begin{tcolorbox}[colback=green!15, colframe=black!50, boxrule=0.5pt, arc=2mm, left=2mm, right=2mm, top=1mm, bottom=3mm, breakable]
\textbf{\textcolor{red}{Question 1.}} Hai mẫu số liệu ghép nhóm $M_1$, $M_2$ có bảng tần số ghép nhóm như sau:
\[
M_1 \quad
\begin{array}{|c|c|c|c|c|c|}
\hline
\text{Nhóm} & [8;10) & [10;12) & [12;14) & [14;16) & [16;18) \\
\hline
\text{Tần số} & 3 & 4 & 8 & 6 & 4 \\
\hline
\end{array}
\]
\[
M_2 \quad
\begin{array}{|c|c|c|c|c|c|}
\hline
\text{Nhóm} & [8;10) & [10;12) & [12;14) & [14;16) & [16;18) \\
\hline
\text{Tần số} & 6 & 8 & 16 & 12 & 8 \\
\hline
\end{array}
\]
Gọi $s_1$, $s_2$ lần lượt là độ lệch chuẩn của mẫu số liệu ghép nhóm $M_1$, $M_2$. Phát biểu nào sau đây là đúng?
\begin{multicols}{4}
\begin{enumerate}[label=\Alph*.]
    \item $s_1 = s_2$.
    \item $s_1 = 2s_2$.
    \item $2s_1 = s_2$.
    \item $4s_1 = s_2$.
\end{enumerate}
\end{multicols}

\begin{tcolorbox}[colback=green!10, colframe=black!50, boxrule=0.5pt, arc=2mm, left=2mm, right=2mm, top=1mm, bottom=3mm, breakable]
\textbf{\textcolor{red}{Solution 1.}} 
\[
M_1 \quad
\begin{array}{|c|c|c|c|c|c|}
\hline
\text{Nhóm} & [8;10) & [10;12) & [12;14) & [14;16) & [16;18) \\
\hline
\text{Tần số} & 3 & 4 & 8 & 6 & 4 \\
\hline
\end{array}
\]
Với $n = 25$, \[
\bar{x} = \dfrac{3 \cdot 9 + 4 \cdot 11 + 8 \cdot 13 + 6 \cdot 15 + 4 \cdot 17}{25} 
= \dfrac{333}{25}.
\]
Độ lệch chuẩn của một mẫu số liệu ghép nhóm được tính theo công thức:
\[
s = \sqrt{\dfrac{3(9 - \bar{x})^2 + 4(11 - \bar{x})^2 + 8(13 - \bar{x})^2 + 6(15 - \bar{x})^2 + 4(17 - \bar{x})^2}{25}} = 2{,}445.
\]
\[
M_2 \quad
\begin{array}{|c|c|c|c|c|c|}
\hline
\text{Nhóm} & [8;10) & [10;12) & [12;14) & [14;16) & [16;18) \\
\hline
\text{Tần số} & 6 & 8 & 16 & 12 & 8 \\
\hline
\end{array}
\]
Với $n = 50$,
\[
\bar{x} = \dfrac{6 \cdot 9 + 8 \cdot 11 + 16 \cdot 13 + 12 \cdot 15 + 8 \cdot 17}{50} 
= \dfrac{333}{25}.
\]
Độ lệch chuẩn của một mẫu số liệu ghép nhóm được tính theo công thức:
\[
s = \sqrt{\dfrac{6(9 - \bar{x})^2 + 8(11 - \bar{x})^2 + 16(13 - \bar{x})^2 + 12(15 - \bar{x})^2 + 8(17 - \bar{x})^2}{50}} = 2{,}445.
\]
\textbf{Lưu ý}: Tần số của các nhóm trong hai mẫu số liệu tương ứng tỉ lệ nên độ lệch chuẩn của chúng bằng nhau.\\
Chọn đáp án \color{red}{\textbf{A}}.
\end{tcolorbox}

\textbf{\textcolor{red}{Explanation 1.}} \\
\textbf{1. Nội dung câu hỏi trong YCCĐ CT2018:} Các số đặc trưng của mẫu số liệu ghép nhóm. \\
\textbf{2. Năng lực đặc thù:} Giải quyết vấn đề toán học. \\
\textbf{3. Tiêu chí biểu hiện thành phần năng lực:} 
GQ2. Lựa chọn, đề xuất được cách thức, giải pháp giải quyết vấn đề. \\
\textbf{4. Chỉ báo biểu hiện năng lực:} 
GQ2.1. Lựa chọn được cách thức, quy trình giải quyết vấn đề. \\
\textbf{5. Cấp độ tư duy:} Thông hiểu. \\
\textbf{6. Dạng thức câu hỏi:} Câu hỏi này thuộc câu trắc nghiệm (TN) nhiều phương án lựa chọn. \\
\textbf{7. Định hướng ôn tập:} Học sinh cần nắm vững phần lý thuyết và cách giải bài toán về Các số đặc trưng của mẫu số liệu ghép nhóm như: \\
- Giải thích được ý nghĩa và vai trò của các số đặc trưng đo mức độ phân tán cho mẫu số liệu ghép nhóm: khoảng biến thiên, khoảng tứ phân vị, phương sai, độ lệch chuẩn trong thực tiễn. \\
- Chỉ ra được những kết luận nhờ ý nghĩa của các số đặc trưng đo mức độ phân tán cho mẫu số liệu ghép nhóm: khoảng biến thiên, khoảng tứ phân vị, phương sai, độ lệch chuẩn trong trường hợp đơn giản. \\
- Tính được các số đặc trưng đo mức độ phân tán cho mẫu số liệu ghép nhóm: khoảng biến thiên, khoảng tứ phân vị, phương sai, độ lệch chuẩn. \\
\textbf{8. Các dạng toán cần ôn tập:} Thành thạo và nắm vững các công thức về Các số đặc trưng đo xu thế trung tâm của mẫu số liệu ghép nhóm và các số đặc trưng đo mức độ phân tán cho mẫu số liệu ghép nhóm như: \\
- Hiểu được ý nghĩa và vai trò của các số đặc trưng nói trên của mẫu số liệu trong thực tiễn. \\
- Tính được các số đặc trưng đo xu thế trung tâm cho mẫu số liệu ghép nhóm: số trung bình cộng (hay số trung bình), trung vị (median), tứ phân vị (quartiles), mốt (mode). \\
- Rút ra được kết luận nhờ ý nghĩa của các số đặc trưng đo xu thế trung tâm nói trên của mẫu số liệu trong trường hợp đơn giản. \\
- Giải thích được ý nghĩa và vai trò của các số đặc trưng đo mức độ phân tán cho mẫu số liệu ghép nhóm: khoảng biến thiên, khoảng tứ phân vị, phương sai, độ lệch chuẩn trong thực tiễn.\\
- Chỉ ra được những kết luận nhờ ý nghĩa của các số đặc trưng đo mức độ phân tán cho mẫu số liệu ghép nhóm: khoảng biến thiên, khoảng tứ phân vị, phương sai, độ lệch chuẩn trong trường hợp đơn giản. \\
- Tính được các số đặc trưng đo mức độ phân tán cho mẫu số liệu ghép nhóm: khoảng biến thiên, khoảng tứ phân vị, phương sai, độ lệch chuẩn. 

\textbf{\textcolor{red}{JSON Data Format Question 1.}} 
\jsonfile{example_assets/example1.json}
\end{tcolorbox}

\noindent
\begin{tcolorbox}[colback=green!15, colframe=black!50, boxrule=0.5pt, arc=2mm, left=2mm, right=2mm, top=1mm, bottom=3mm, breakable]
    \begin{minipage}[t]{0.78\textwidth}
        \vspace{0pt}
        \textbf{\textcolor{red}{Question 2.}} Cho hàm số 
        $y = \dfrac{ax + b}{cx + d}, \ (c \neq 0,\ ad - bc \neq 0)$ 
        có đồ thị như hình vẽ bên. Tiệm cận ngang của đồ thị hàm số là
        \begin{multicols}{2}
            \begin{enumerate}[label=\Alph*.]
                \item $x = -1$.
                \item $y = \dfrac{1}{2}$.
                \item $y = -1$.
                \item $x = \dfrac{1}{2}$.
            \end{enumerate}
        \end{multicols}
    \end{minipage}
    \hfill
    \begin{minipage}[t]{0.2\textwidth}
        \vspace{5pt}
        \centering
        \includegraphics[width=\linewidth]{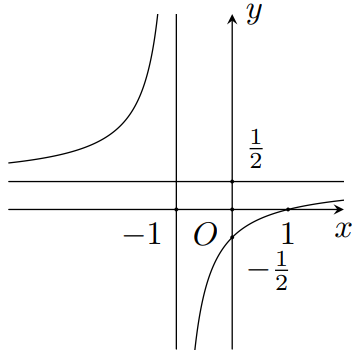}
    \end{minipage}

\begin{tcolorbox}[colback=green!10, colframe=black!50, boxrule=0.5pt, arc=2mm, left=2mm, right=2mm, top=1mm, bottom=3mm, breakable]
\textbf{\textcolor{red}{Solution 2.}} 
Dựa vào đồ thị ta thấy 
\[
\lim_{x \to +\infty} y = \dfrac{1}{2} 
\quad \text{và} \quad 
\lim_{x \to -\infty} y = \dfrac{1}{2}.
\]
Do đó, tiệm cận ngang của đồ thị hàm số là \( y = \dfrac{1}{2} \).\\
Chọn đáp án \color{red}{\textbf{B}}.
\end{tcolorbox}

\textbf{\textcolor{red}{Explanation 2.}} \\
\textbf{1. Nội dung câu hỏi trong YCCĐ CT 2018:} Khảo sát và vẽ đồ thị của hàm số. \\
\textbf{2. Năng lực đặc thù:} Tư duy và lập luận toán học. \\
\textbf{3. Tiêu chí biểu hiện năng lực:} 
TD1. Thực hiện được các thao tác tư duy như: so sánh, phân tích, tổng hợp, đặc biệt hóa, khái quát hóa, tương tự; quy nạp, diễn dịch. \\
\textbf{4. Chỉ báo biểu hiện năng lực:} 
TD1.3. Lí giải được kết quả của việc quan sát. \\
\textbf{5. Cấp độ tư duy:} Nhận biết. \\
\textbf{6. Dạng thức câu hỏi:} Câu hỏi này thuộc câu trắc nghiệm (TN) nhiều phương án lựa chọn. \\
\textbf{7. Định hướng ôn tập:} Học sinh cần nắm vững phần lý thuyết và cách giải bài toán về Khảo sát và vẽ đồ thị của hàm số như: \\
- Nhận biết được hình ảnh hình học của đường tiệm cận ngang, đường tiệm cận đứng, đường tiệm cận xiên của đồ thị hàm số. \\
- Nhận biết được tính đối xứng (trục đối xứng, tâm đối xứng) của đồ thị các hàm số. \\
- Mô tả được sơ đồ tổng quát để khảo sát hàm số (tìm tập xác định, xét chiều biến thiên, tìm cực trị, tìm tiệm cận, lập bảng biến thiên, vẽ đồ thị). \\
\textbf{8. Các dạng toán cần ôn tập:} Thành thạo các dạng toán về Khảo sát và vẽ đồ thị của hàm số như:\\
- Nhận biết được hình ảnh hình học của đường tiệm cận ngang, đường tiệm cận đứng, đường tiệm cận xiên của đồ thị hàm số.\\
- Nhận biết được tính đối xứng (trục đối xứng, tâm đối xứng) của đồ thị các hàm số.\\
- Mô tả được sơ đồ tổng quát để khảo sát hàm số (tìm tập xác định, xét chiều biến thiên, tìm cực trị, tìm tiệm cận, lập bảng biến thiên, vẽ đồ thị).

\textbf{\textcolor{red}{JSON Data Format Question 2.}} 
\jsonfile{example_assets/example2.json}
\end{tcolorbox}

\subsection{Sample Questions – Section II}

\noindent
\begin{tcolorbox}[colback=green!15, colframe=black!50, boxrule=0.5pt, arc=2mm, left=2mm, right=2mm, top=1mm, bottom=3mm, breakable]
    \textbf{\textcolor{red}{Question 3.}} Các thiên thạch có đường kính lớn hơn 140 m và có thể lại gần Trái Đất ở khoảng cách nhỏ hơn 7.500.000 km được coi là những vật thể có khả năng va chạm gây nguy hiểm cho Trái Đất. Để theo dõi những thiên thạch này, người ta đã thiết lập các trạm quan sát các vật thể bay gần Trái Đất. Giả sử có một hệ thống quan sát có khả năng theo dõi các vật thể ở độ cao không vượt quá 6.600 km so với mực nước biển. Coi Trái Đất là khối cầu có bán kính 6.400 km. Chọn hệ trục tọa độ $Oxyz$ trong không gian có gốc $O$ tại tâm Trái Đất và đơn vị độ dài trên mỗi trục tọa độ là 1.000 km. Một thiên thạch (coi như một hạt) chuyển động với tốc độ không đổi theo đường thẳng từ điểm $M(6; 20; 0)$ đến điểm $N(-6; -12; 16)$. \\
    a) Đường thẳng $MN$ có phương trình tham số là $\begin{cases}
    x = 6 + 3t \\
    y = 20 + 8t \\
    z = -4t
    \end{cases}
    \quad (t \in \mathbb{R}).$ \\
    b) Vị trí đầu tiên thiên thạch di chuyển vào phạm vi theo dõi các trạm quan sát là điểm $A(-3;-4;12)$. \\
    c) Khoảng cách giữa vị trí đầu tiên và vị trí cuối cùng mà thiên thạch di chuyển trong phạm vi theo dõi của hệ thống quan sát là 18.900 km (kết quả làm tròn đến hàng trăm theo đơn vị ki-lô-mét). \\ 
    d) Nếu thời gian di chuyển của thiên thạch trong phạm vi theo dõi các trạm quan sát là $3$ phút thì thời gian từ di chuyển từ $M$ đến $N$ là $6$ phút. 
\begin{center}
\includegraphics[scale=0.3]{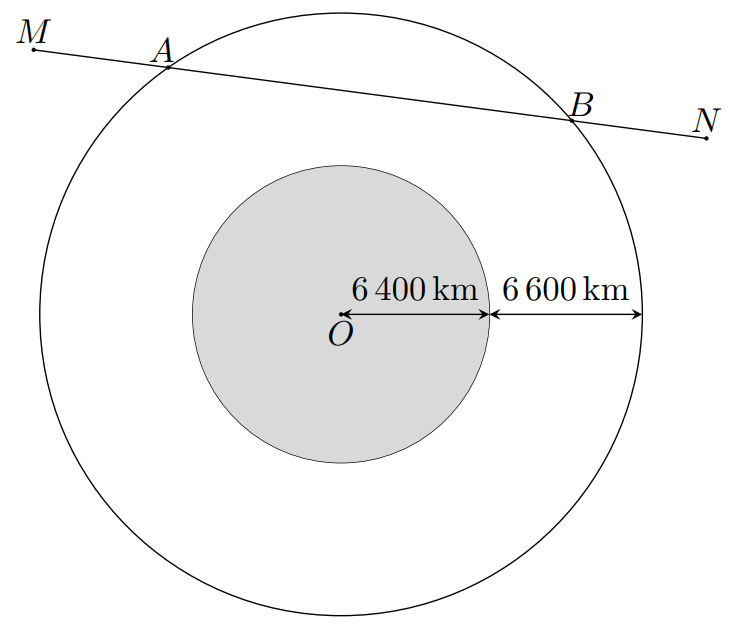}   
\end{center}
\end{tcolorbox}
    
\begin{tcolorbox}[colback=green!10, colframe=black!50, boxrule=0.5pt, arc=2mm, left=2mm, right=2mm, top=1mm, bottom=3mm, breakable]
\textbf{\textcolor{red}{Solution 3.}}\\
a) Đúng.\\
Ta có $M(6; 20; 0)$, $N(-6; -12; 16)$; suy ra $\overrightarrow{MN} = (-12; -32; 16)$.\\
Chọn $\vec{u} = (3; 8; -4)$ làm một vector chỉ phương của đường thẳng $MN$.\\
Khi đó, đường thẳng $MN$ có phương trình tham số là
\[
\begin{cases}
x = 6 + 3t \\
y = 20 + 8t \quad (t \in \mathbb{R}) \\
z = -4t
\end{cases}
\]
b) Sai.\\
Phạm vi theo đồ thị là mặt cầu (O): \( x^2 + y^2 + z^2 = 13^2 \).\\
Tọa độ giao điểm của \( MN \) với (O) là nghiệm của phương trình
\[
(6 + 3t)^2 + (20 + 8t)^2 + (-4t)^2 = 13^2
\]
\[
\Leftrightarrow 89t^2 + 35t + 267 = 0
\]
\[
\Leftrightarrow
\begin{cases}
t = -1 \\
t = -3
\end{cases}
\]
Với \( t = -1 \) ta có \( A(3; 12; 4) \).\\
Với \( t = -3 \) ta có \( B(-3; -4; 12) \).\\
Tọa độ \( \overrightarrow{MA} = (-3; -8; 4) \), \( \overrightarrow{MB} = (-9; -24; 12) \); suy ra \( \overrightarrow{MB} = 3\overrightarrow{MA} \).\\
Vậy điểm giữa tia \( A(3; 12; 4) \).\\
c) Đúng.\\
Vì \( AB = \sqrt{(-3 - 3)^2 + (-4 - 12)^2 + (12 - 4)^2} = \sqrt{356} \) và đổi ví dụ dài trên mỗi trục là 1000 km nên khoảng cách \( AB \simeq 18900 \, \text{km}. \)\\
d) Đúng.\\
Vì \( AB = 2\sqrt{89}, MN = 4\sqrt{89} \) nên \( t_{MN} = 2 \cdot t_{AB} = 2 \cdot 3 = 6 \, (\text{phút}). \)\\
Chọn đáp án: a) đúng | b) sai | c) đúng | d) đúng.
\end{tcolorbox}

\begin{tcolorbox}[colback=green!10, colframe=black!50, boxrule=0.5pt, arc=2mm, left=2mm, right=2mm, top=1mm, bottom=3mm, breakable]
\textbf{\textcolor{red}{Explanation 3.}} \\
\textbf{1. Nội dung câu hỏi trong YCCĐ CT 2018:} Phương trình đường thẳng trong không gian và Phương trình mặt cầu. \\
\textbf{2. Năng lực đặc thù (Câu 4a):} Tư duy và lập luận toán học. \\
\textbf{3. Tiêu chí biểu hiện thành phần năng lực:} \\
TD1. Thực hiện được các thao tác tư duy như: so sánh, phân tích, tổng hợp, đặc biệt hóa, khái quát hóa, tương tự; quy nạp, diễn dịch. \\
\textbf{4. Chỉ báo biểu hiện năng lực:} \\
TD1.2. Phát hiện được sự tương đồng và khác biệt trong những tình huống tương đối phức tạp. \\
\textbf{5. Cấp độ tư duy:} Thông hiểu. \\
\textbf{6. Năng lực đặc thù (Câu 4b):} Mô hình hóa toán học. \\
\textbf{7. Tiêu chí biểu hiện thành phần năng lực:} \\
MH2. Giải quyết được những vấn đề toán học trong mô hình được thiết lập. \\
\textbf{8. Chỉ báo biểu hiện năng lực:} \\
MH2.1. Giải quyết được những vấn đề toán học trong mô hình được thiết lập. \\
\textbf{9. Cấp độ tư duy:} Thông hiểu. \\
\textbf{10. Năng lực đặc thù (Câu 4c):} Mô hình hóa toán học. \\
\textbf{11. Tiêu chí biểu hiện thành phần năng lực:} \\
MH2. Giải quyết được những vấn đề toán học trong mô hình được thiết lập. \\
\textbf{12. Chỉ báo biểu hiện năng lực:} \\
MH2.1. Giải quyết được những vấn đề toán học trong mô hình được thiết lập. \\
\textbf{13. Cấp độ tư duy:} Thông hiểu. \\
\textbf{14. Năng lực đặc thù (Câu 4d):} Mô hình hóa toán học. \\
\textbf{15. Tiêu chí biểu hiện thành phần năng lực:} \\
MH2. Giải quyết được những vấn đề toán học trong mô hình được thiết lập. \\
\textbf{16. Chỉ báo biểu hiện năng lực:} \\
MH2.1. Giải quyết được những vấn đề toán học trong mô hình được thiết lập. \\
\textbf{17. Cấp độ tư duy:} Vận dụng. \\
\textbf{18. Dạng thức câu hỏi:} Câu hỏi này thuộc câu trắc nghiệm (TN) Đúng-Sai. \\
\textbf{19. Định hướng ôn tập:} Học sinh cần nắm vững phần lý thuyết và cách giải bài toán về Phương trình đường thẳng trong không gian và Phương trình mặt cầu như: \\
- Nhận biết được phương trình tổng quát của mặt phẳng. \\
- Thiết lập được phương trình tổng quát của mặt phẳng trong hệ trục tọa độ $Oxyz$ theo một trong ba cách cơ bản: qua một điểm và biết vectơ pháp tuyến; qua một điểm và biết cặp vectơ chỉ phương (suy ra vectơ pháp tuyến nhờ vào việc tìm vectơ vuông góc với cặp vectơ chỉ phương); qua ba điểm không thẳng hàng. \\
- Thiết lập được điều kiện để hai mặt phẳng song song, vuông góc với nhau. \\
- Tính được khoảng cách từ một điểm đến một mặt phẳng bằng phương pháp tọa độ. \\
- Vận dụng được kiến thức về phương trình đường thẳng trong không gian để giải một số bài toán liên quan đến thực tiễn. \\
- Nhận biết được phương trình mặt cầu. \\
- Xác định được tâm, bán kính của mặt cầu khi biết phương trình của nó. \\
- Thiết lập được phương trình của mặt cầu khi biết tâm và bán kính. \\
- Vận dụng được kiến thức về phương trình mặt cầu để giải một số bài toán liên quan đến thực tiễn. \\
\textbf{20. Các dạng toán cần ôn tập:} Thành thạo các dạng toán về Phương trình mặt phẳng; Phương trình đường thẳng trong không gian và Phương trình mặt cầu như: \\
- Nhận biết được phương trình tổng quát của mặt phẳng. \\
- Thiết lập được phương trình tổng quát của mặt phẳng trong hệ trục tọa độ $Oxyz$ theo một trong ba cách cơ bản: qua một điểm và biết vectơ pháp tuyến; qua một điểm và biết cặp vectơ chỉ phương (suy ra vectơ pháp tuyến nhờ vào việc tìm vectơ vuông góc với cặp vectơ chỉ phương); qua ba điểm không thẳng hàng. \\
- Thiết lập được điều kiện để hai mặt phẳng song song, vuông góc với nhau. \\
- Tính được khoảng cách từ một điểm đến một mặt phẳng bằng phương pháp tọa độ. \\
- Vận dụng được kiến thức về phương trình mặt phẳng để giải một số bài toán liên quan đến thực tiễn. \\
- Nhận biết được phương trình chính tắc, phương trình tham số, vectơ chỉ phương của đường thẳng trong không gian. \\
- Thiết lập được phương trình chính tắc, phương trình tham số của đường thẳng: qua một điểm và biết một vectơ chỉ phương; qua hai điểm. \\
- Xác định được điều kiện để hai đường thẳng chéo nhau, cắt nhau, song song hoặc vuông góc với nhau. \\
- Thiết lập được công thức tính góc giữa hai đường thẳng, góc giữa đường thẳng và mặt phẳng; góc giữa hai mặt phẳng. \\
- Vận dụng được kiến thức về phương trình đường thẳng trong không gian để giải một số bài toán liên quan đến thực tiễn. \\
- Nhận biết được phương trình mặt cầu. \\
- Xác định được tâm, bán kính của mặt cầu khi biết phương trình của nó. \\
- Thiết lập được phương trình của mặt cầu khi biết tâm và bán kính. \\
- Vận dụng được kiến thức về phương trình mặt cầu để giải một số bài toán liên quan đến thực tiễn.
\end{tcolorbox}

\noindent
\textbf{\textcolor{red}{JSON Data Format Question 3.}} 
\jsonfile{example_assets/example3.json}

\subsection{Sample Questions – Section III}

\noindent
\begin{tcolorbox}[colback=green!15, colframe=black!50, boxrule=0.5pt, arc=2mm, left=2mm, right=2mm, top=1mm, bottom=3mm, breakable]
    \begin{minipage}[t]{0.65\textwidth}
        \vspace{0pt}
        \textbf{\textcolor{red}{Question 4.}} Kiến trúc sư thiết kế một khu sinh hoạt cộng đồng có dạng hình chữ nhật với chiều rộng và chiều dài lần lượt là 60 m và 80 m. Trong đó, phần được tô màu đậm là sân chơi, phần còn lại để trồng hoa. Mỗi phần trồng hoa có đường biên cong là một phần của parabol với đỉnh thuộc một trục đối xứng của hình chữ nhật và khoảng cách từ đỉnh đó đến trung điểm cạnh tương ứng của hình chữ nhật bằng 20 m (xem hình minh họa). Diện tích của phần sân chơi là bao nhiêu mét vuông?
    \end{minipage}
    \hfill
    \begin{minipage}[t]{0.3\textwidth}
        \vspace{0pt}
        \centering
        \includegraphics[width=\linewidth]{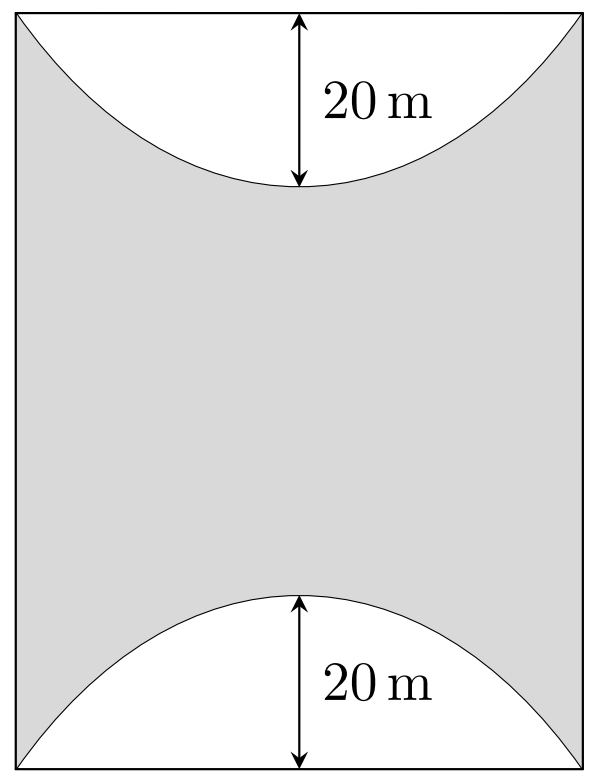}
    \end{minipage}
\end{tcolorbox}

\begin{tcolorbox}[colback=green!10, colframe=black!50, boxrule=0.5pt, arc=2mm, left=2mm, right=2mm, top=1mm, bottom=3mm, breakable]
\textbf{\textcolor{red}{Solution 4.}}\\
\begin{center}
\includegraphics[scale=0.3]{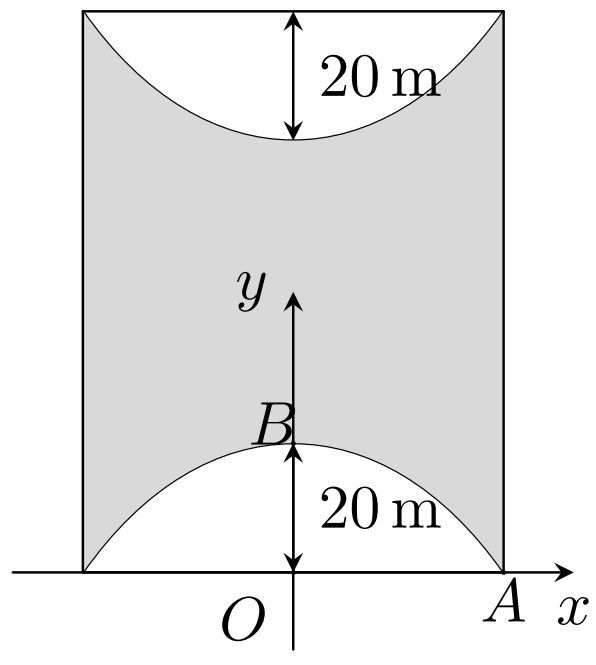}\\
\end{center}
Chọn hệ trục Oxy sao cho \( A(30; 0), B(0; 20) \). Khi đó, parabol \( (P) \) có đỉnh là \( B(0; 20) \) và đi qua điểm \( A(30; 0) \), nên \( (P) \) có phương trình là
\[
y = -\dfrac{1}{45}x^2 + 20.
\]
Khi đó, tổng diện tích các phần parabol là \( 4 \int_0^{30} \left( -\dfrac{1}{45}x^2 + 20 \right) dx = 1600 \, (\text{m}^2) \). Vậy diện tích phần sân chơi là \( 60 \times 80 - 1600 = 3200 \, (\text{m}^2) \).\\
Chọn đáp án \color{red}{\textbf{B}}.
\end{tcolorbox}

\begin{tcolorbox}[colback=green!15, colframe=black!50, boxrule=0.5pt, arc=2mm, left=2mm, right=2mm, top=1mm, bottom=3mm, breakable]
\textbf{\textcolor{red}{Explanation 4.}} \\
\textbf{1. Nội dung câu hỏi trong YCCĐ CT 2018:} Tích phân. Ứng dụng hình học của tích phân. \\
\textbf{2. Năng lực đặc thù:} Mô hình hóa toán học. \\
\textbf{3. Tiêu chí biểu hiện năng lực:} MH2. Giải quyết được những vấn đề toán học trong mô hình được thiết lập. \\
\textbf{4. Chỉ báo biểu hiện năng lực:} MH2.1. Giải quyết được những vấn đề toán học trong mô hình được thiết lập. \\
\textbf{5. Cấp độ tư duy:} Vận dụng. \\
\textbf{6. Dạng thức câu hỏi:} Câu hỏi này thuộc câu trắc nghiệm (TN) trả lời ngắn. \\
\textbf{7. Định hướng ôn tập:} Học sinh cần nắm vững phần lý thuyết và cách giải bài toán về Tích phân. Ứng dụng hình học của tích phân như: \\
- Nhận biết được định nghĩa và các tính chất của tích phân. \\
- Tính được tích phân trong những trường hợp đơn giản. \\
- Sử dụng được tích phân để tính diện tích của một số hình phẳng, thể tích của một số hình khối. \\
- Vận dụng được tích phân để giải một số bài toán có liên quan đến thực tiễn. \\
\textbf{8. Các dạng toán cần ôn tập:} Thành thạo các dạng toán về Tích phân. Ứng dụng hình học của tích phân như: \\
- Nhận biết được định nghĩa và các tính chất của tích phân. \\
- Tính được tích phân trong những trường hợp đơn giản. \\
- Sử dụng được tích phân để tính diện tích của một số hình phẳng, thể tích của một số hình khối. \\
- Vận dụng được tích phân để giải một số bài toán có liên quan đến thực tiễn. \\
- Ngoài ra cần nắm vững các dạng toán về nguyên hàm và các bài toán thực tế về nguyên hàm.
\end{tcolorbox}
\noindent
\textbf{\textcolor{red}{JSON Data Format Question 4.}} 
\jsonfile{example_assets/example4.json}

\section{Examples of creative exam question generation}
\label{sec: AppendixB}
\setcounter{equation}{0}

\begin{tcolorbox}[colback=yellow!15, colframe=black!50, boxrule=0.5pt, arc=2mm, left=2mm, right=2mm, top=1mm, bottom=3mm, breakable]
\textbf{\textcolor{red}{Question Generation. ID = [8\_III\_3]}} \\
Công ty $X$ có hai xí nghiệp I và II cùng sản xuất một loại sản phẩm.  Xí nghiệp I sản xuất $60\%$ tổng sản phẩm và có tỉ lệ phế phẩm là $4\%$, xí nghiệp II sản xuất $40\%$ tổng sản phẩm và có tỉ lệ phế phẩm là $3\%$. Người ta dùng một đồng xu cân đối, đồng chất để gieo ngẫu nhiên. Nếu ra mặt ngửa thì chọn một sản phẩm của xí nghiệp I, nếu ra mặt sấp thì chọn một sản phẩm của cả hai xí nghiệp. Tính xác suất để chọn được phế phẩm, biết rằng đó là phế phẩm của xí nghiệp I (làm tròn đến hàng phần trăm).
\end{tcolorbox}

\begin{tcolorbox}[colback=yellow!15, colframe=black!50, boxrule=0.5pt, arc=2mm, left=2mm, right=2mm, top=1mm, bottom=3mm, breakable]
\textbf{\textcolor{red}{Question Generation. ID = [8\_III\_3]}} \\
Có hai chiếc hộp, hộp I có $6$ quả bóng màu đỏ và $4$ quả bóng màu vàng, hộp II có $7$ quả bóng màu đỏ và $3$ quả bóng màu vàng, các quả bóng có cùng kích thước và khối lượng. Lấy ngẫu nhiên một quả bóng từ hộp I bỏ vào hộp II. Sau đó, lấy ra ngẫu nhiên một quả bóng từ hộp II. Tính xác suất để quả bóng được lấy ra từ hộp II là quả bóng được chuyển từ hộp I sang, biết rằng quả bóng đó có màu đỏ (làm tròn kết quả đến hàng phần trăm).
\end{tcolorbox}

\begin{tcolorbox}[colback=yellow!15, colframe=black!50, boxrule=0.5pt, arc=2mm, left=2mm, right=2mm, top=1mm, bottom=3mm, breakable]
\textbf{\textcolor{red}{Question Generation. ID = [8\_III\_3]}} \\
Hộp thứ nhất có 4 viên bi xanh và 6 viên bi vàng. Hộp thứ hai có 5 viên bi xanh và 4 viên bi vàng. Các viên bi có cùng kích thước và khối lượng. Lấy ra ngẫu nhiên 1 viên bi từ hộp thứ nhất chuyển sang hộp thứ hai. Sau đó lại lấy ra ngẫu nhiên 1 viên bi từ hộp thứ hai. Biết rằng viên bi đó có màu xanh, tính xác suất để lấy được viên bi từ hộp thứ hai cũng là viên bi được chuyển sang từ hộp thứ nhất (làm tròn kết quả đến hàng phần trăm).
\end{tcolorbox}

\begin{tcolorbox}[colback=blue!5, colframe=black!50, boxrule=0.5pt, arc=2mm, left=2mm, right=2mm, top=1mm, bottom=3mm, breakable]
\textbf{\textcolor{red}{Question Generation. ID = [11\_III\_3]}} \\
Trong không gian $Oxyz$, đơn vị trên mỗi trục là $2000\,\text{km}$, các kỹ sư hàng không vũ trụ mô phỏng bề mặt Mặt Trăng dưới dạng mặt cầu $(S): x^2 + y^2 + z^2 - 2x - 2y - 2z = 0$. Một vệ tinh truyền tín hiệu được gửi đến bởi tàu thăm dò từ Trái Đất đang ở vị trí $A(2;2;0)$. Để truyền tín hiệu tốt nhất, các nhà khoa học cần đặt một trạm chuyển tiếp tại vị trí $B$ sao cho $B$ nằm trên bề mặt Mặt Trăng, thỏa mãn hoành độ điểm $B$ là số dương, tam giác $OAB$ là tam giác vuông cân tại $B$. Tìm khoảng cách thực tế từ trạm chuyển tiếp $B$ đến vị trí tiếp nhận tín hiệu trên Mặt Trăng ở $C(0;2;0)$, là nơi hạ cánh của tàu vũ trụ. (Làm tròn đến hàng phần trăm của nghìn km).
\end{tcolorbox}

\begin{tcolorbox}[colback=blue!5, colframe=black!50, boxrule=0.5pt, arc=2mm, left=2mm, right=2mm, top=1mm, bottom=3mm, breakable]
\textbf{\textcolor{red}{Question Generation. ID = [11\_III\_3]}} \\
Một khu rừng rộng lớn có hai tháp phát tín hiệu vô tuyến đặt tại hai vị trí $A(10;6;-2)$ và $B(5;10;-9)$. Toàn bộ khu vực rừng nằm trên một mặt đồi có dạng phẳng, được mô tả bởi mặt phẳng $(\alpha): 2x + 2y + z - 12 = 0$ (đơn vị trên mỗi trục tọa độ là km). Một thiết bị thu sóng $M$ cần được đặt tại một điểm bất kỳ trên mặt đồi $(\alpha)$ để nhận tín hiệu tốt nhất từ cả hai tháp $A$ và $B$ sao cho đường truyền tín hiệu từ $A$ đến $M$ và từ $B$ đến $M$ phải tạo các góc bằng nhau với mặt phẳng đồi $(\alpha)$. Thiết bị $M$ sẽ luôn nằm trên một đường tròn cố định có bán kính là bao nhiêu km?
\end{tcolorbox}

\begin{tcolorbox}[colback=blue!5, colframe=black!50, boxrule=0.5pt, arc=2mm, left=2mm, right=2mm, top=1mm, bottom=3mm, breakable]
\textbf{\textcolor{red}{Question Generation. ID = [11\_III\_3]}} \\
ROV (Remotely Operated Vehicle) là phương tiện điều khiển từ xa dưới nước, được sử dụng rộng rãi trong nghiên cứu biển, khảo sát dầu khí, và tìm kiếm cứu hộ. Một ROV đang ở vị trí $M\left(\tfrac{1}{2}; \tfrac{\sqrt{5}}{2}; 0\right)$ trong không gian $Oxyz$ (đơn vị: km), với tâm $O(0;0;0)$ là trạm định vị trung tâm dưới đáy biển. ROV được giao nhiệm vụ khảo sát một khu vực hình cầu dưới biển được mô tả bởi phương trình $x^2 + y^2 + z^2 = 8$. ROV di chuyển theo một đường thẳng và các cảm biến của nó có thể thu thập dữ liệu từ hai điểm $A$ và $B$ nằm trên bề mặt của khu vực mặt cầu. Để tối ưu hóa việc thu thập dữ liệu và đạt được phạm vi khảo sát hiệu quả nhất, đội ngũ vận hành cần xác định hướng di chuyển của ROV sao cho diện tích của tam giác $OAB$ lớn nhất là bao nhiêu $\text{km}^2$? (Làm tròn kết quả đến hàng phần trăm).
\end{tcolorbox}

\end{appendix}
\end{document}

%% file: supports/initialize_command.tex
\usepackage{lineno}
\usepackage{xspace}
\usepackage{times}
\usepackage{enumitem}
\usepackage{amsfonts,amsmath,amssymb,amsthm,amsopn}
\usepackage[dvipsnames]{xcolor}
\usepackage{bm}
\usepackage{graphicx,graphics,color}
\usepackage{epsfig}
\usepackage{caption,subcaption}
\usepackage{array}
\usepackage{multirow}
\usepackage{longtable}
\usepackage{threeparttable}
\usepackage{dcolumn}
\newcolumntype{d}[1]{D{.}{.}{#1}}
\usepackage{hyperref}

\usepackage{natbib}
\usepackage[capitalise,sort&compress]{cleveref}
\usepackage[toc]{appendix}
\usepackage{appendix}

\newcolumntype{M}[1]{>{\centering\arraybackslash}m{#1}}

\crefname{figure}{Fig.}{Figs.}
\crefname{table}{Table}{Tables}
\crefname{equation}{Eq.}{Eqs.}

\crefname{section}{Section}{Sections}
\crefname{subsection}{Subsection}{Subsections}
\crefname{chapter}{Chapter}{Chapters}
\crefname{appendix}{}{}

\usepackage{bigints}

\newcommand{\blue}[1]{\textcolor{blue}{#1}}

\newcommand{\bv}{\begin{verbatim}}
\newcommand{\V}{\verb}                  % EX: \V=-d{#@~}= Expr must




%% file: Manuscript.bbl
\begin{thebibliography}{24}
\expandafter\ifx\csname natexlab\endcsname\relax\def\natexlab#1{#1}\fi
\providecommand{\url}[1]{\texttt{#1}}
\providecommand{\href}[2]{#2}
\providecommand{\path}[1]{#1}
\providecommand{\DOIprefix}{doi:}
\providecommand{\ArXivprefix}{arXiv:}
\providecommand{\URLprefix}{URL: }
\providecommand{\Pubmedprefix}{pmid:}
\providecommand{\doi}[1]{\href{http://dx.doi.org/#1}{\path{#1}}}
\providecommand{\Pubmed}[1]{\href{pmid:#1}{\path{#1}}}
\providecommand{\bibinfo}[2]{#2}
\ifx\xfnm\relax \def\xfnm[#1]{\unskip,\space#1}\fi
\bibitem[{Dao et~al.(2023{\natexlab{a}})Dao, Le, Vo, Phan, Ngo, Nguyen, Nguyen, and Nguyen}]{dao2023vnhsgevietnamesehighschool}
\bibinfo{author}{X.-Q. Dao}, \bibinfo{author}{N.-B. Le}, \bibinfo{author}{T.-D. Vo}, \bibinfo{author}{X.-D. Phan}, \bibinfo{author}{B.-B. Ngo}, \bibinfo{author}{V.-T. Nguyen}, \bibinfo{author}{T.-M.-T. Nguyen}, \bibinfo{author}{H.-P. Nguyen}, \bibinfo{title}{Vnhsge: Vietnamese high school graduation examination dataset for large language models}, \bibinfo{year}{2023}{\natexlab{a}}. \URLprefix \url{https://arxiv.org/abs/2305.12199}. \href{http://arxiv.org/abs/2305.12199}{{\tt arXiv:2305.12199}}.
\bibitem[{Dao et~al.(2023{\natexlab{b}})Dao, Le, Phan, and Ngo}]{dao2023chatgptpassvietnamesenational}
\bibinfo{author}{X.-Q. Dao}, \bibinfo{author}{N.-B. Le}, \bibinfo{author}{X.-D. Phan}, \bibinfo{author}{B.-B. Ngo}, \bibinfo{title}{Can chatgpt pass the vietnamese national high school graduation examination?}, \bibinfo{year}{2023}{\natexlab{b}}. \URLprefix \url{https://arxiv.org/abs/2306.09170}. \href{http://arxiv.org/abs/2306.09170}{{\tt arXiv:2306.09170}}.
\bibitem[{Dao and Le(2023)}]{dao2023chatgptgoodbingchat}
\bibinfo{author}{X.-Q. Dao}, \bibinfo{author}{N.-B. Le}, \bibinfo{title}{Chatgpt is good but bing chat is better for vietnamese students}, \bibinfo{year}{2023}. \URLprefix \url{https://arxiv.org/abs/2307.08272}. \href{http://arxiv.org/abs/2307.08272}{{\tt arXiv:2307.08272}}.
\bibitem[{Wang et~al.(2024)Wang, Wang, Zhu, Wang, Tran, and Du}]{WANG2024124167}
\bibinfo{author}{S.~Wang}, \bibinfo{author}{F.~Wang}, \bibinfo{author}{Z.~Zhu}, \bibinfo{author}{J.~Wang}, \bibinfo{author}{T.~Tran}, \bibinfo{author}{Z.~Du},
\newblock \bibinfo{title}{Artificial intelligence in education: A systematic literature review},
\newblock \bibinfo{journal}{Expert Systems with Applications} \bibinfo{volume}{252} (\bibinfo{year}{2024}) \bibinfo{pages}{124167}. \DOIprefix\doi{https://doi.org/10.1016/j.eswa.2024.124167}.
\bibitem[{Awang et~al.(2025)Awang, Yusop, and Danaee}]{Liz2025mathematicseducation}
\bibinfo{author}{L.~A. Awang}, \bibinfo{author}{F.~D. Yusop}, \bibinfo{author}{M.~Danaee},
\newblock \bibinfo{title}{Current practices and future direction of artificial intelligence in mathematics education: A systematic review},
\newblock \bibinfo{journal}{International Electronic Journal of Mathematics Education} \bibinfo{volume}{20} (\bibinfo{year}{2025}). \DOIprefix\doi{https://doi.org/10.29333/iejme/16006}.
\bibitem[{Liang et~al.(2025)Liang, Zhang, Zhong, and Liu}]{liang2025mathematicsmachinecreativitysurvey}
\bibinfo{author}{S.~Liang}, \bibinfo{author}{W.~Zhang}, \bibinfo{author}{T.~Zhong}, \bibinfo{author}{T.~Liu}, \bibinfo{title}{Mathematics and machine creativity: A survey on bridging mathematics with ai}, \bibinfo{year}{2025}. \URLprefix \url{https://arxiv.org/abs/2412.16543}. \href{http://arxiv.org/abs/2412.16543}{{\tt arXiv:2412.16543}}.
\bibitem[{Arnau-González et~al.(2025)Arnau-González, Solera-Monforte, Wu, and Arevalillo-Herráez}]{ARNAUGONZALEZ2025126663}
\bibinfo{author}{P.~Arnau-González}, \bibinfo{author}{S.~Solera-Monforte}, \bibinfo{author}{Y.~Wu}, \bibinfo{author}{M.~Arevalillo-Herráez},
\newblock \bibinfo{title}{A framework for adapting conversational intelligent tutoring systems to enable collaborative learning},
\newblock \bibinfo{journal}{Expert Systems with Applications} \bibinfo{volume}{271} (\bibinfo{year}{2025}) \bibinfo{pages}{126663}. \DOIprefix\doi{https://doi.org/10.1016/j.eswa.2025.126663}.
\bibitem[{Nancy et~al.(2024)Nancy, Stefania, and Andrew}]{nancy2024benchmarkmathmisconceptionsbridging}
\bibinfo{author}{O.~Nancy}, \bibinfo{author}{D.~Stefania}, \bibinfo{author}{L.~Andrew}, \bibinfo{title}{A benchmark for math misconceptions: Bridging gaps in middle school algebra with ai-supported instruction}, \bibinfo{year}{2024}. \URLprefix \url{https://arxiv.org/abs/2412.03765}. \href{http://arxiv.org/abs/2412.03765}{{\tt arXiv:2412.03765}}.
\bibitem[{Zhao et~al.(2025)Zhao, Han, Wu, He, Ning, Yuan, Li, Wang, and Song}]{ZHAO2025113905}
\bibinfo{author}{D.~Zhao}, \bibinfo{author}{D.~Han}, \bibinfo{author}{J.~Wu}, \bibinfo{author}{Z.~He}, \bibinfo{author}{B.~Ning}, \bibinfo{author}{Y.~Yuan}, \bibinfo{author}{Y.~Li}, \bibinfo{author}{C.~Wang}, \bibinfo{author}{S.~Song},
\newblock \bibinfo{title}{Enhancing math reasoning ability of large language models via computation logic graphs},
\newblock \bibinfo{journal}{Knowledge-Based Systems} \bibinfo{volume}{325} (\bibinfo{year}{2025}) \bibinfo{pages}{113905}. \DOIprefix\doi{https://doi.org/10.1016/j.knosys.2025.113905}.
\bibitem[{Li et~al.(2026)Li, Wang, Jose, and Ge}]{LI2026103577}
\bibinfo{author}{L.~Li}, \bibinfo{author}{Z.~Wang}, \bibinfo{author}{J.~M. Jose}, \bibinfo{author}{X.~Ge},
\newblock \bibinfo{title}{Llm supporting knowledge tracing leveraging global subject and student specific knowledge graphs},
\newblock \bibinfo{journal}{Information Fusion} \bibinfo{volume}{126} (\bibinfo{year}{2026}) \bibinfo{pages}{103577}. \DOIprefix\doi{https://doi.org/10.1016/j.inffus.2025.103577}.
\bibitem[{Di and JoyJiaoW(2025)}]{di2025enhancingmathreasoningsmallsized}
\bibinfo{author}{X.~Di}, \bibinfo{author}{JoyJiaoW}, \bibinfo{title}{Enhancing math reasoning in small-sized llms via preview difficulty-aware intervention}, \bibinfo{year}{2025}. \URLprefix \url{https://arxiv.org/abs/2508.01604}. \href{http://arxiv.org/abs/2508.01604}{{\tt arXiv:2508.01604}}.
\bibitem[{Shah et~al.(2025)Shah, Yu, Lyu, Park, Yu, He, Ke, Mozer, Bengio, Arora, and Goyal}]{shah2025aiassistedgenerationdifficultmath}
\bibinfo{author}{V.~Shah}, \bibinfo{author}{D.~Yu}, \bibinfo{author}{K.~Lyu}, \bibinfo{author}{S.~Park}, \bibinfo{author}{J.~Yu}, \bibinfo{author}{Y.~He}, \bibinfo{author}{N.~R. Ke}, \bibinfo{author}{M.~Mozer}, \bibinfo{author}{Y.~Bengio}, \bibinfo{author}{S.~Arora}, \bibinfo{author}{A.~Goyal}, \bibinfo{title}{Ai-assisted generation of difficult math questions}, \bibinfo{year}{2025}. \URLprefix \url{https://arxiv.org/abs/2407.21009}. \href{http://arxiv.org/abs/2407.21009}{{\tt arXiv:2407.21009}}.
\bibitem[{Oh(2025)}]{10817549}
\bibinfo{author}{S.~Oh},
\newblock \bibinfo{title}{Evaluating mathematical problem-solving abilities of generative ai models: Performance analysis of o1-preview and gpt-4o using the korean college scholastic ability test},
\newblock \bibinfo{journal}{IEEE Access} \bibinfo{volume}{13} (\bibinfo{year}{2025}) \bibinfo{pages}{1227--1235}. \DOIprefix\doi{10.1109/ACCESS.2024.3523703}.
\bibitem[{Frieder et~al.(2023)Frieder, Pinchetti, Chevalier, Griffiths, Salvatori, Lukasiewicz, Petersen, and Berner}]{frieder2023mathematicalcapabilitieschatgpt}
\bibinfo{author}{S.~Frieder}, \bibinfo{author}{L.~Pinchetti}, \bibinfo{author}{A.~Chevalier}, \bibinfo{author}{R.-R. Griffiths}, \bibinfo{author}{T.~Salvatori}, \bibinfo{author}{T.~Lukasiewicz}, \bibinfo{author}{P.~C. Petersen}, \bibinfo{author}{J.~Berner}, \bibinfo{title}{Mathematical capabilities of chatgpt}, \bibinfo{year}{2023}. \URLprefix \url{https://arxiv.org/abs/2301.13867}. \href{http://arxiv.org/abs/2301.13867}{{\tt arXiv:2301.13867}}.
\bibitem[{Trinh et~al.(2024)Trinh, Wu, Le, He, and Luong}]{AlphaGeometryTrinh2024}
\bibinfo{author}{T.~Trinh}, \bibinfo{author}{Y.~Wu}, \bibinfo{author}{Q.~Le}, \bibinfo{author}{H.~He}, \bibinfo{author}{T.~Luong},
\newblock \bibinfo{title}{Solving olympiad geometry without human demonstrations},
\newblock \bibinfo{journal}{Nature}  (\bibinfo{year}{2024}). \DOIprefix\doi{10.1038/s41586-023-06747-5}.
\bibitem[{Chervonyi et~al.(2025)Chervonyi, Trinh, Olšák, Yang, Nguyen, Menegali, Jung, Verma, Le, and Luong}]{chervonyi2025goldmedalistperformancesolvingolympiad}
\bibinfo{author}{Y.~Chervonyi}, \bibinfo{author}{T.~H. Trinh}, \bibinfo{author}{M.~Olšák}, \bibinfo{author}{X.~Yang}, \bibinfo{author}{H.~Nguyen}, \bibinfo{author}{M.~Menegali}, \bibinfo{author}{J.~Jung}, \bibinfo{author}{V.~Verma}, \bibinfo{author}{Q.~V. Le}, \bibinfo{author}{T.~Luong}, \bibinfo{title}{Gold-medalist performance in solving olympiad geometry with alphageometry2}, \bibinfo{year}{2025}. \URLprefix \url{https://arxiv.org/abs/2502.03544}. \href{http://arxiv.org/abs/2502.03544}{{\tt arXiv:2502.03544}}.
\bibitem[{Liang et~al.(2025)Liang, Song, Li, Yang, Zhang, Mi, and Yu}]{liang2025solvingchallengingimoproblems}
\bibinfo{author}{Z.~Liang}, \bibinfo{author}{L.~Song}, \bibinfo{author}{Y.~Li}, \bibinfo{author}{T.~Yang}, \bibinfo{author}{F.~Zhang}, \bibinfo{author}{H.~Mi}, \bibinfo{author}{D.~Yu}, \bibinfo{title}{Towards solving more challenging imo problems via decoupled reasoning and proving}, \bibinfo{year}{2025}. \URLprefix \url{https://arxiv.org/abs/2507.06804}. \href{http://arxiv.org/abs/2507.06804}{{\tt arXiv:2507.06804}}.
\bibitem[{Huang and Yang(2025)}]{huang2025gemini25procapable}
\bibinfo{author}{Y.~Huang}, \bibinfo{author}{L.~F. Yang}, \bibinfo{title}{Gemini 2.5 pro capable of winning gold at imo 2025}, \bibinfo{year}{2025}. \URLprefix \url{https://arxiv.org/abs/2507.15855}. \href{http://arxiv.org/abs/2507.15855}{{\tt arXiv:2507.15855}}.
\bibitem[{Dao and Le(2023)}]{dao2023investigatingeffectivenesschatgptmathematical}
\bibinfo{author}{X.-Q. Dao}, \bibinfo{author}{N.-B. Le}, \bibinfo{title}{Investigating the effectiveness of chatgpt in mathematical reasoning and problem solving: Evidence from the vietnamese national high school graduation examination}, \bibinfo{year}{2023}. \URLprefix \url{https://arxiv.org/abs/2306.06331}. \href{http://arxiv.org/abs/2306.06331}{{\tt arXiv:2306.06331}}.
\bibitem[{Dil et~al.(2025)Dil, Chen, and Damevski}]{DIL2025112581}
\bibinfo{author}{C.~Dil}, \bibinfo{author}{H.~Chen}, \bibinfo{author}{K.~Damevski},
\newblock \bibinfo{title}{Towards higher quality software vulnerability data using llm-based patch filtering},
\newblock \bibinfo{journal}{Journal of Systems and Software} \bibinfo{volume}{230} (\bibinfo{year}{2025}) \bibinfo{pages}{112581}. \DOIprefix\doi{https://doi.org/10.1016/j.jss.2025.112581}.
\bibitem[{Zhou et~al.(2025)Zhou, Chen, Guo, Yan, Lee, Wang, Lee, Zhang, Shao, Yang, and Wang}]{zhou2025mementofinetuningllmagents}
\bibinfo{author}{H.~Zhou}, \bibinfo{author}{Y.~Chen}, \bibinfo{author}{S.~Guo}, \bibinfo{author}{X.~Yan}, \bibinfo{author}{K.~H. Lee}, \bibinfo{author}{Z.~Wang}, \bibinfo{author}{K.~Y. Lee}, \bibinfo{author}{G.~Zhang}, \bibinfo{author}{K.~Shao}, \bibinfo{author}{L.~Yang}, \bibinfo{author}{J.~Wang}, \bibinfo{title}{Memento: Fine-tuning llm agents without fine-tuning llms}, \bibinfo{year}{2025}. \URLprefix \url{https://arxiv.org/abs/2508.16153}. \href{http://arxiv.org/abs/2508.16153}{{\tt arXiv:2508.16153}}.
\bibitem[{Hatalis et~al.(2025)Hatalis, Christou, and Kondapalli}]{hatalis2025reviewcasebasedreasoningllm}
\bibinfo{author}{K.~Hatalis}, \bibinfo{author}{D.~Christou}, \bibinfo{author}{V.~Kondapalli}, \bibinfo{title}{Review of case-based reasoning for llm agents: Theoretical foundations, architectural components, and cognitive integration}, \bibinfo{year}{2025}. \URLprefix \url{https://arxiv.org/abs/2504.06943}. \href{http://arxiv.org/abs/2504.06943}{{\tt arXiv:2504.06943}}.
\bibitem[{Zhao et~al.(2024)Zhao, Kang, Wang, and He}]{zhao2024doclayoutyoloenhancingdocumentlayout}
\bibinfo{author}{Z.~Zhao}, \bibinfo{author}{H.~Kang}, \bibinfo{author}{B.~Wang}, \bibinfo{author}{C.~He}, \bibinfo{title}{Doclayout-yolo: Enhancing document layout analysis through diverse synthetic data and global-to-local adaptive perception}, \bibinfo{year}{2024}. \URLprefix \url{https://arxiv.org/abs/2410.12628}. \href{http://arxiv.org/abs/2410.12628}{{\tt arXiv:2410.12628}}.
\bibitem[{Memarian and Doleck(2024)}]{MEMARIAN2024100040}
\bibinfo{author}{B.~Memarian}, \bibinfo{author}{T.~Doleck},
\newblock \bibinfo{title}{A review of assessment for learning with artificial intelligence},
\newblock \bibinfo{journal}{Computers in Human Behavior: Artificial Humans} \bibinfo{volume}{2} (\bibinfo{year}{2024}) \bibinfo{pages}{100040}. \DOIprefix\doi{https://doi.org/10.1016/j.chbah.2023.100040}.

\end{thebibliography}
